\documentclass[journal]{IEEEtran}
\usepackage{cite}
\usepackage{bm}
\usepackage{flushend}
\usepackage{stfloats}
\usepackage{multicol}
\usepackage{multirow}
\usepackage{array}
\usepackage{graphicx}
\usepackage{epstopdf}
\usepackage{setspace}
\usepackage{lettrine}
\usepackage{amsmath}
\usepackage{amsbsy}
\usepackage{mathrsfs}
\usepackage{amsfonts,amssymb}
\usepackage{algorithm}
\usepackage{algorithmic}
\usepackage{graphicx}
\usepackage{subfigure}
\usepackage{booktabs}
\usepackage{amsmath} 
\usepackage{amssymb}  

\begin{document}

%
\title{ \huge{Robust Learning with Kernel Mean $p$-Power Error Loss}}

\author{Badong Chen, \emph{Senior Member, IEEE}, Lei Xing, \emph{Student Member, IEEE}, Xin Wang, \emph{Student Member, IEEE},\\ Jing Qin, \emph{Member, IEEE},
 Nanning Zheng, \emph{Fellow, IEEE}}

\maketitle

\begin{abstract}
Correntropy is a second order statistical measure in kernel space, which has been successfully applied in robust learning and signal processing. In this paper, we define a non-second order statistical measure in kernel space, called the \emph{kernel mean-p power error} (KMPE), including the \emph{correntropic loss} (C-Loss) as a special case. Some basic properties of KMPE are presented. In particular, we apply the KMPE to \emph{extreme learning machine} (ELM) and \emph{principal component analysis} (PCA), and develop two robust learning algorithms, namely ELM-KMPE and PCA-KMPE. Experimental results on synthetic and benchmark data show that the developed algorithms can achieve consistently better performance when compared with some existing methods.

\end{abstract}


\let\thefootnote\relax\footnotetext{This work was supported by 973 Program (No. 2015CB351703) and National NSF of China (No. 91648208, No. 61372152).
\par Badong Chen, Lei Xing, Xin Wang and Nanning Zheng are with the Institute of Artificial Intelligence and Robotics, Xi'an Jiaotong University, Xi'an, 710049, China.({chenbd; nnzheng }@mail.xjtu.edu.cn; {xl2010; wangxin0420}@stu.xjtu.edu.cn).
\par Jing Qin is with the Center of Smart Health, School of Nursing, The Hong Kong Polytechnic University, Hongkong, China.(harryqinjingcn@gmail.com).
}


%
\IEEEpeerreviewmaketitle
\section{Introduction}
\lettrine[lines=2]{T}{HE} basic framework in learning theory generally considers learning from examples by optimizing (minimizing or maximizing) a certain loss function such that the learned model can discover the structures (or dependencies) in the data generating system under the uncertainty caused by noise or unknown knowledge about the system {\cite{principe2010information}}. The second order statistical measures such as mean square error (MSE), variance and correlation and have been commonly used as the loss functions in machine learning or adaptive system training due to their simplicity and mathematical tractability. For example, the goal of the \emph{least squares} (LS) regression is to learn an unknown mapping (linear or nonlinear) such that MSE between the model output and desired response is minimized. Also, the orthogonal linear transformation in \emph{principal component analysis} (PCA) is determined such that the first principal component has the largest possible variance, and each succeeding component in turn has the highest variance possible under the constraint that it is orthogonal to the preceding components {\cite{jolliffe2002principal}}. The \emph{canonical-correlation analysis} (CCA) is another example, where the goal is to find the linear combinations of the components in two random vectors which have maximum correlation with each other {\cite{hardoon2004canonical}}.
\par The loss functions based on the second order statistical measures, however, are sensitive to outliers in the data, and are not good solution to learning with non-Gaussian data in general {\cite{principe2010information}}. To handle non-Gaussian data (or noises), various non-second order (or non-quadratic) loss functions are frequently applied to learning systems. Typical examples include Huber's min-max loss {\cite{rousseeuw2005robust,black1998robust}}, Lorentzian error loss {\cite{black1998robust}}, risk-sensitive loss {\cite{boel2002robustness}} and mean p-power error (MPE) loss {\cite{pei1994least,chen2015smoothed}}. The MPE is the $p$-th absolute moment of the error, which with a proper $p$ value can deal with non-Gaussian data well. In general, MPE is robust to large outliers when $p < 2$ {\cite{pei1994least}}. Information theoretic measures, such as entropy, KL divergence and mutual information can also be used as loss functions in machine learning and non-Gaussian signal processing since they can capture higher order statistics (i.e. moments or correlations beyond second order) of the data {\cite{principe2010information}}. Many numerical examples have shown the superior performance of \emph{information theoretic learning}(ITL) {\cite{principe2010information,chen2013system}}. Particularly in recent years, a novel ITL similarity measure, called correntropy, has been successfully applied to robust learning and signal processing {\cite{liu2007correntropy,he2014robust,he2011robust,chen2017maximum,chen2016generalized,chen2015convergence,chen2014steady,mandanas2016robust,zhu2016correntropy}}. Correntropy is a generalized correlation in high dimensional kernel space (usually induced by a Gaussian kernel), which is directly related to the probability of how similar two random variables are in a neighborhood (controlled by the kernel bandwidth) of the joint space {\cite{liu2007correntropy}}. Since correntropy is a local similarity measure, it can increase the robustness with respect to outliers by assigning small weights to data beyond the neighborhood.
\par Essentially, correntropy is a second order statistical measure (i.e. correlation) in kernel space, which corresponds to a non-second order measure in original space. Similarly, one can define other second order statistical measures, such as MSE, in kernel space. The MSE in kernel space is also called the \emph{correntropic loss} (C-Loss) {\cite{singh2014c,chen2016efficient}}. It can be shown that minimizing the C-Loss is equivalent to maximizing the correntropy. In this paper, we define a non-second order measure in kernel space, called \emph{kernel mean p-power error} (KMPE), which is the MPE in kernel space and, of course, is also a non-second order measure in original space. The KMPE will reduce to the C-Loss as $p=2$, but with a proper $p$ value can outperform the C-Loss when used as a loss function in robust learning. In the present work, we focus mainly on two application examples, \emph{extreme learning machine} (ELM) {\cite{huang2006extreme,huang2011extreme}} and PCA. The ELM is a single-hidden-layer feedforward neural network (SLFN) with randomly generated hidden nodes, which can be used for regression, classification and many other learning tasks {\cite{huang2006extreme,huang2011extreme}}. The proposed KMPE will be used to develop robust ELM and PCA algorithms.
\par The rest of the paper is structured as follows. In section II, we define the KMPE, and give some basic properties. In section III, we apply the KMPE to ELM and PCA, and develop the ELM-KMPE and PCA-KMPE algorithms. In section IV, we present experimental results to demonstrate the desirable performance of the new algorithms. Finally in section V, we give the conclusion.

\section{KERNEL MEAN P-POWER ERROR}
\subsection{Definition}
Non-second order statistical measures can be defined elegantly as a second order measure in kernel space. For example, the correntropy between two random variables $X$ and $Y$, is a correlation measure in kernel space, given by {\cite{liu2007correntropy}}
\begin{equation}
\begin{aligned}
V(X,Y) &= \textbf{E}\left[ {{{\left\langle {\Phi (X),\Phi (Y)} \right\rangle }_{\mathcal{H}}}} \right] \\
&= \int {{{\left\langle {\Phi (x),\Phi (y)} \right\rangle }_\mathcal{H}}d{F_{XY}}(x,y)}
\end{aligned}
\end{equation}
where $\textbf{E}[.]$ denotes the expectation operator, ${F_{XY}}(x,y)$ stands for the joint distribution function, and $\Phi (x) = \kappa (x,.)$ is a nonlinear mapping induced by a Mercer kernel $\kappa (.,.)$, which transforms $x$ from the original space to a functional Hilbert space (or kernel space) $\mathcal{H}$ equipped with an inner product ${\left\langle {.,.} \right\rangle _\mathcal{H}}$ satisfying ${\left\langle {\Phi (x),\Phi (y)} \right\rangle _\mathcal{H}} = \kappa (x,y)$. Obviously, we have $V(X,Y) = \textbf{E}\left[ {\kappa (X,Y)} \right]$. In this paper, without mentioned otherwise, the kernel function is a Gaussian kernel, given by
\begin{equation}
\kappa (x,y) = {\kappa _\sigma }(x - y) = \exp \left( { - \frac{{{{(x - y)}^2}}}{{2{\sigma ^2}}}} \right)
\end{equation}
with $\sigma$ being the kernel bandwidth. Similarly, the C-Loss as MSE in kernel space, can be defined by {\cite{chen2016generalized}}
\begin{equation}
\begin{aligned}
&C(X,Y) = \frac{1}{2}\textbf{E}\left[ {\left\| {\Phi (X) - \Phi (Y)} \right\|_\mathcal{H}^2} \right]\\
&{\rm{              = }}\frac{1}{2}\textbf{E}\left[ {{{\left\langle {\Phi (X) - \Phi (Y),\Phi (X) - \Phi (Y)} \right\rangle }_\mathcal{H}}} \right]\\
&{\rm{              = }}\frac{1}{2}\textbf{E}\left[ {{{\left\langle {\Phi (X),\Phi (X)} \right\rangle }_\mathcal{H}}{\rm{ + }}{{\left\langle {\Phi (Y),\Phi (Y)} \right\rangle }_\mathcal{H}} - 2{{\left\langle {\Phi (X),\Phi (Y)} \right\rangle }_\mathcal{H}}} \right]\\
&{\rm{              = }}\frac{1}{2}\textbf{E}\left[ {2{\kappa _\sigma }(0) - 2{\kappa _\sigma }(X - Y)} \right]\\
&{\rm{             =}} \textbf{E}\left[ {1 - {\kappa _\sigma }(X - Y)} \right]
\end{aligned}
\end{equation}
where $1/2$ is inserted to make the expression more convenient. It holds that $C(X,Y) = 1 - V(X,Y)$, hence minimizing the C-Loss will be equivalent to maximizing the correntropy. The \emph{maximum correntropy criterion} (MCC) has drawn more and more attention recently due to its robustness to large outliers {\cite{liu2007correntropy,he2014robust,he2011robust,chen2017maximum,chen2016generalized,chen2015convergence,chen2014steady,mandanas2016robust,zhu2016correntropy}}.

\par In this work, we define a new statistical measure in kernel space in a non-second order manner. Specifically, we generalize the C-Loss to the case of arbitrary power and define the mean p-power error (MPE) in kernel space, and call the new measure the kernel MPE (KMPE). Given two random variables $X$ and $Y$, the KMPE is defined by
\begin{equation}
\begin{aligned}
{C_p}(X,Y) &= {2^{ - {p \mathord{\left/
 {\vphantom {p 2}} \right.
 \kern-\nulldelimiterspace} 2}}}\textbf{E}\left[ {\left\| {\Phi (X) - \Phi (Y)} \right\|_\mathcal{H}^p} \right]\\
&{\rm{              }} = {2^{ - {p \mathord{\left/
 {\vphantom {p 2}} \right.
 \kern-\nulldelimiterspace} 2}}}\textbf{E}\left[ {{{\left( {\left\| {\Phi (X) - \Phi (Y)} \right\|_\mathcal{H}^2} \right)}^{{p \mathord{\left/
 {\vphantom {p 2}} \right.
 \kern-\nulldelimiterspace} 2}}}} \right]\\
&{\rm{              }} = {2^{ - {p \mathord{\left/
 {\vphantom {p 2}} \right.
 \kern-\nulldelimiterspace} 2}}}\textbf{E}\left[ {{{\left( {2 - 2{\kappa _\sigma }(X - Y)} \right)}^{{p \mathord{\left/
 {\vphantom {p 2}} \right.
 \kern-\nulldelimiterspace} 2}}}} \right]\\
&{\rm{              }} = \textbf{E}\left[ {{{\left( {1 - {\kappa _\sigma }(X - Y)} \right)}^{{p \mathord{\left/
 {\vphantom {p 2}} \right.
 \kern-\nulldelimiterspace} 2}}}} \right]
\end{aligned}
\end{equation}
where $p>0$ is the power parameter. Clearly, the KMPE includes the C-Loss as a special case (when $p=2$ ). In addition, given $N$ samples $\left\{ {{x_i},{y_i}} \right\}_{i = 1}^N$ , the \emph{empirical KMPE} can be easily obtained as
\begin{equation}
{\hat C_p}(X,Y) = \frac{1}{N}\sum\limits_{i = 1}^N {{{\left( {1 - {\kappa _\sigma }({x_i} - {y_i})} \right)}^{{p \mathord{\left/
 {\vphantom {p 2}} \right.
 \kern-\nulldelimiterspace} 2}}}}
\end{equation}
Since ${\hat C_p}(X,Y)$ is a function of the sample vectors $\textbf{X} = {\left[ {{x_1},{x_2}, \cdots ,{x_N}} \right]^T}$ and $\textbf{Y} = {\left[ {{y_1},{y_2}, \cdots ,{y_N}} \right]^T}$, one can also denote ${\hat C_p}(X,Y)$ by ${\hat C_p}(\textbf{X},\textbf{Y})$ if no confusion arises.

\subsection{Properties}
Some basic properties of the proposed KMPE are presented below.
\par \emph{Property 1}: ${C_p}(X,Y)$ is symmetric, that is ${C_p}(X,Y)={C_p}(Y,X)$.
\par \emph{Proof}: Straightforward since ${\kappa _\sigma }(X - Y) = {\kappa _\sigma }(Y - X)$.
\par \emph{Property 2}: ${C_p}(X,Y)$ is positive and bounded: $0 \le {C_p}(X,Y) < 1$, and it reaches its minimum if and only if $X = Y$.
\par \emph{Proof}: Straightforward since $0 < {\kappa _\sigma }(X - Y) \le 1$, with ${\kappa _\sigma }(X - Y) = 1$ if and only if $X=Y$.
\par \emph{Property 3}: As $p$ is small enough, it holds that ${C_p}(X,Y) \approx 1{\rm{ + }}\frac{p}{2}\textbf{E}\left[ {\log \left( {1 - {\kappa _\sigma }(X - Y)} \right)} \right]$.
\par \emph{Proof}: The property holds since ${\left( {1 - {\kappa _\sigma }(X - Y)} \right)^{{p \mathord{\left/
 {\vphantom {p 2}} \right.
 \kern-\nulldelimiterspace} 2}}} \approx 1{\rm{ + }}\frac{p}{2}\log \left( {1 - {\kappa _\sigma }(X - Y)} \right)$ for $p$ small enough.
\par \emph{Property 4}: As $\sigma$ is large enough, it holds that ${C_p}(X,Y) \approx {\left( {2{\sigma ^2}} \right)^{{{ - p} \mathord{\left/
 {\vphantom {{ - p} 2}} \right.
 \kern-\nulldelimiterspace} 2}}}\textbf{E}\left[ {{{\left| {X - Y} \right|}^p}} \right]$.
\par \emph{Proof}: Since $\exp (x) \approx 1 + x$ for $x$ small enough, as $\sigma  \to \infty $, we have
\begin{equation}
\begin{aligned}
{\left( {1 - {\kappa _\sigma }(X - Y)} \right)^{{p \mathord{\left/
 {\vphantom {p 2}} \right.
 \kern-\nulldelimiterspace} 2}}} &= {\left( {1 - \exp \left( { - \frac{{{{\left( {X - Y} \right)}^2}}}{{2{\sigma ^2}}}} \right)} \right)^{{p \mathord{\left/
 {\vphantom {p 2}} \right.
 \kern-\nulldelimiterspace} 2}}}\\
&{\rm{                            }} \approx {\left( {\frac{{{{\left( {X - Y} \right)}^2}}}{{2{\sigma ^2}}}} \right)^{{p \mathord{\left/
 {\vphantom {p 2}} \right.
 \kern-\nulldelimiterspace} 2}}}\\
 &= {\left( {2{\sigma ^2}} \right)^{{{ - p} \mathord{\left/
 {\vphantom {{ - p} 2}} \right.
 \kern-\nulldelimiterspace} 2}}}{\left| {X - Y} \right|^p}
\end{aligned}
\end{equation}
\par \emph{Remark}: By Property 4, one can conclude that the KMPE will be, approximately, equivalent to the MPE when kernel bandwidth $\sigma$ is large enough.
\par \emph{Property 5}: Let $\emph{\textbf{e}} = \textbf{X} - \textbf{Y} = {\left[ {{e_1},{e_2}, \cdots ,{e_N}} \right]^T}$, where ${e_i} = {x_i} - {y_i}$. if $p \ge 2$, the empirical KMPE ${\hat C_p}(\textbf{X},\textbf{Y})$ as a function of $\emph{\textbf{e}}$ is convex at any point satisfying
${\left\| \emph{\textbf{e}} \right\|_\infty } = \mathop {\max }\limits_{i = 1,2, \cdots ,N} \left| {{e_i}} \right| \le \sigma $.
\par \emph{Proof}: Since ${\hat C_p}(\textbf{X},\textbf{Y}) = \frac{1}{N}\sum\limits_{i = 1}^N {{{\left( {1 - {\kappa _\sigma }({e_i})} \right)}^{{p \mathord{\left/
 {\vphantom {p 2}} \right.
 \kern-\nulldelimiterspace} 2}}}} $, the Hessian matrix of ${\hat C_p}(\textbf{X},\textbf{Y})$ with respect to $\emph{\textbf{e}}$ is
\begin{equation}
{H_{{{\hat C}_p}(\textbf{X},\textbf{Y})}}\left( \emph{\textbf{e}} \right) = \left[ {\frac{{{\partial ^2}{{\hat C}_p}(\textbf{X},\textbf{Y})}}{{\partial {e_i}\partial {e_j}}}} \right] = diag\left[ {{\xi _1},{\xi _2}, \cdots ,{\xi _N}} \right]
\end{equation}
where
\begin{equation}
\begin{aligned}
{\xi _i} &= \frac{p}{{4N{\sigma ^4}}}{\left( {1 - {\kappa _\sigma }({e_i})} \right)^{{{(p - 4)} \mathord{\left/
 {\vphantom {{(p - 4)} 2}} \right.
 \kern-\nulldelimiterspace} 2}}}{\kappa _\sigma }({e_i}) \times \\
&{\rm{             }}\left\{ {(p - 2)e_i^2{\kappa _\sigma }({e_i}) - 2e_i^2\left( {1 - {\kappa _\sigma }({e_i})} \right) + 2{\sigma ^2}\left( {1 - {\kappa _\sigma }({e_i})} \right)} \right\}
\end{aligned}
\end{equation}
When $p \ge 2$, we have ${\xi _i} \ge 0$ if $\left| {{e_i}} \right| \le \sigma $. Thus, for any point $\emph{\textbf{e}}$ with ${\left\| \emph{\textbf{e}} \right\|_\infty } \le \sigma $ , we have ${H_{{{\hat C}_p}(\textbf{X},\textbf{Y})}}\left( \emph{\textbf{e}} \right) \ge 0$.

\par \emph{Property 6}: Given any point $\emph{\textbf{e}}$ with ${\left\| \emph{\textbf{e}} \right\|_\infty } > \sigma $, the empirical KMPE ${\hat C_p}(X,Y)$ will be convex at $\emph{\textbf{e}}$ if $p$ is larger than a certain value.
\par \emph{Proof}: From (8), if $\left| {{e_i}} \right| \le \sigma $ and $p\ge2$, or if $\left| {{e_i}} \right| > \sigma $ and $p \ge \frac{{2\left[ {e_i^2 - {\sigma ^2}} \right]\left( {1 - {\kappa _\sigma }({e_i})} \right)}}{{e_i^2{\kappa _\sigma }({e_i})}} + 2$, we have ${\xi _i} \ge 0$. So, it holds that ${H_{{{\hat C}_p}(\textbf{X},\textbf{Y})}}\left( \emph{\textbf{e}} \right) \ge 0$ if
\begin{equation}
p  \ge \mathop {\max }\limits_{i= 1, \cdots ,N\atop| {e_i} | > \sigma }\left \{{\frac{{2\left[ {e_i^2 - {\sigma ^2}} \right]\left( {1 - {\kappa _\sigma }({e_i})} \right)}}{{e_i^2{\kappa _\sigma }({e_i})}} + 2}\right \}
\end{equation}
This complete the proof.

\par \emph{Remark}: According to Property 5 and 6, the empirical KMPE as a function of \emph{\textbf{e}} is convex at any point with ${\left\| \emph{\textbf{e}} \right\|_\infty } \le \sigma $. and it can also be convex at a point with ${\left\| \emph{\textbf{e}} \right\|_\infty } > \sigma $ if the power parameter $p$ is larger than a certain value.

\par \emph{Property 7}: Let $\textbf{0}$ be an $N$-dimensional zero vector. Then as $\sigma  \to \infty $ (or ${x_i} \to 0,i = 1, \cdots ,N$), it holds that
\begin{equation}
{\hat C_p}(\textbf{X},\textbf{0}) \approx \frac{1}{{N{{\left( {\sqrt 2 \sigma } \right)}^p}}}\left\| \textbf{X} \right\|_p^p
\end{equation}
where $\left\| \textbf{X} \right\|_p^p = \sum\limits_{i = 1}^N {{{\left| {{x_i}} \right|}^p}} $.

\par \emph{Proof}: As $\sigma$ is large enough, we have
\begin{equation}
\begin{aligned}
{{\hat C}_p}(\textbf{X},\textbf{0}) &= \frac{1}{N}\sum\limits_{i = 1}^N {{{\left( {1 - {\kappa _\sigma }({x_i})} \right)}^{{p \mathord{\left/
 {\vphantom {p 2}} \right.
 \kern-\nulldelimiterspace} 2}}}} \\
&{\rm{             }}\mathop  \approx \limits^{(a)} \frac{1}{N}\sum\limits_{i = 1}^N {{{\left( {1 - \left( {1 - \frac{{x_i^2}}{{2{\sigma ^2}}}} \right)} \right)}^{{p \mathord{\left/
 {\vphantom {p 2}} \right.
 \kern-\nulldelimiterspace} 2}}}} \\
&{\rm{             }} = \frac{1}{N}\sum\limits_{i = 1}^N {{{\left( {\frac{{x_i^2}}{{2{\sigma ^2}}}} \right)}^{{p \mathord{\left/
 {\vphantom {p 2}} \right.
 \kern-\nulldelimiterspace} 2}}}} \\
&{\rm{             }} = \frac{1}{{N{{\left( {\sqrt 2 \sigma } \right)}^p}}}\sum\limits_{i = 1}^N {{{\left| {{x_i}} \right|}^p}}
\end{aligned}
\end{equation}

\par \emph{Property 8}: Assume that $\left| {{x_i}} \right| > \delta $, $\forall i:{x_i} \ne 0$, where $\delta$ is a small positive number. As  $\sigma  \to 0 + $, minimizing the empirical KMPE ${\hat C_p}(\textbf{X},\textbf{0})$ will be, approximately, equivalent to minimizing the ${l_0}$-norm of $\textbf{X}$, that is
\begin{equation}
\mathop {\min }\limits_{\textbf{X} \in \Omega } {\hat C_p}(\textbf{X},\textbf{0}) \sim \mathop {\min }\limits_{\textbf{X} \in \Omega } {\left\| \textbf{X} \right\|_0},\;\;as\;\; \sigma  \to 0 +
\end{equation}
where $\Omega $ denotes a feasible set of $\textbf{X}$.

\par \emph{Proof}: Let ${\textbf{X}_0}$ be the solution obtained by minimizing ${\left\| \textbf{X} \right\|_0}$ over $\Omega $ and ${ \textbf{X} _C}$ the solution achieved by minimizing ${\hat C_p}(\textbf{X},\textbf{0})$. Then ${\hat C_p}({\textbf{X}_C},\textbf{0}) \le {\hat C_p}({\textbf{X}_0},\textbf{0})$, and

\begin{equation}
\small
\begin{aligned}
&\sum\limits_{i = 1}^N {\left[ {{{\left( {1 - {\kappa _\sigma }\left( {{{({\textbf{X}_C})}_i}} \right)} \right)}^{{p \mathord{\left/
 {\vphantom {p 2}} \right.
 \kern-\nulldelimiterspace} 2}}} - 1} \right]}  \\
 &\le \sum\limits_{i = 1}^N {\left[ {{{\left( {1 - {\kappa _\sigma }\left( {{{({\textbf{X}_0})}_i}} \right)} \right)}^{{p \mathord{\left/
 {\vphantom {p 2}} \right.
 \kern-\nulldelimiterspace} 2}}} - 1} \right]}
\end{aligned}
\end{equation}

where ${({\textbf{X}_C})_i}$ denotes the $i$th component of ${\textbf{X}_C}$. It follows that
\begin{small}
\begin{equation}
\begin{aligned}
&{\left\| {{\textbf{X}_C}} \right\|_0} - N + \sum\limits_{i = 1,{{({\textbf{X}_C})}_i} \ne 0}^N {\left[ {{{\left( {1 - {\kappa _\sigma }\left( {{{({\textbf{X}_C})}_i}} \right)} \right)}^{{p \mathord{\left/
 {\vphantom {p 2}} \right.
 \kern-\nulldelimiterspace} 2}}} - 1} \right]}  \\
 &\le {\left\| {{\textbf{X}_0}} \right\|_0} - N + \sum\limits_{i = 1,{{({\textbf{X}_0})}_i} \ne 0}^N {\left[ {{{\left( {1 - {\kappa _\sigma }\left( {{{({\textbf{X}_0})}_i}} \right)} \right)}^{{p \mathord{\left/
 {\vphantom {p 2}} \right.
 \kern-\nulldelimiterspace} 2}}} - 1} \right]}
\end{aligned}
\end{equation}
\end{small}
Hence
\begin{small}
\begin{equation}
\begin{aligned}
{\left\| {{\textbf{X}_C}} \right\|_0} - {\left\| {{\textbf{X}_0}} \right\|_0} &\le \sum\limits_{i = 1,{{({\textbf{X}_0})}_i} \ne 0}^N {\left[ {{{\left( {1 - {\kappa _\sigma }\left( {{{({\textbf{X}_0})}_i}} \right)} \right)}^{{p \mathord{\left/
 {\vphantom {p 2}} \right.
 \kern-\nulldelimiterspace} 2}}} - 1} \right]}  \\
 &- \sum\limits_{i = 1,{{({\textbf{X}_C})}_i} \ne 0}^N {\left[ {{{\left( {1 - {\kappa _\sigma }\left( {{{({\textbf{X}_C})}_i}} \right)} \right)}^{{p \mathord{\left/
 {\vphantom {p 2}} \right.
 \kern-\nulldelimiterspace} 2}}} - 1} \right]}
\end{aligned}
\end{equation}
\end{small}
Since $\left| {{x_i}} \right| > \delta $, $\forall i:{x_i} \ne 0$ , as $\sigma  \to 0 + $ the right hand side of (15) will approach zero. Thus, if $\sigma$ is small enough, it holds that
\begin{equation}
{\left\| {{\textbf{X}_0}} \right\|_0} \le {\left\| {{\textbf{X}_C}} \right\|_0} \le {\left\| {{\textbf{X}_0}} \right\|_0} + \varepsilon
\end{equation}
where $\varepsilon $ is a small positive number arbitrarily close to zero. This completes the proof.
\par \emph{Remark}: From Property 7 and 8, one can see that the empirical KMPE ${\hat C_p}(\textbf{X},\textbf{0})$ behaves like an $L_p$ norm of $\textbf{X}$ when kernel bandwidth $\sigma$ is very large, and like an $L_0$ norm of $\textbf{X}$ when $\sigma$ is very small.

\section{APPLICATION EXAMPLES}
There are many applications in areas of machine learning and signal processing that can employ the KMPE to solve robustly the relevant problems. In this section, we present two examples to investigate the benefits from the KMPE.
\subsection{Extreme Learning Machine}
The first example is about the Extreme Learning Machine (ELM), a single-hidden-layer feedforward neural network (SLFN) with random hidden nodes {\cite{huang2006extreme,huang2011extreme}}. With a quadratic loss function, the ELM usually requires no iterative tuning and the global optima can be solved in a batch mode. In the following, we use the KMPE as the loss function for ELM, and develop a robust algorithm to train the model. Since there is no closed-form solution under the KMPE loss, the new algorithm will be a fixed-point iterative algorithm.
\par Given $N$ distinct training samples $\{ {\textbf{x}_i},{t_i}\} _{i = 1}^N$, with ${\textbf{x}_i} = {[{x_{i1}},{x_{i2}},...,{x_{id}}]^T} \in {\mathbb{R}^d}$ being the input vector and the ${t_i} \in \mathbb{R}$ target response, the output of a standard SLFN with $L$  hidden nodes will be
\begin{equation}
{y_i} = \sum\limits_{j = 1}^L {{\beta _j}f({\textbf{w}_j} \cdot {\textbf{x}_i} + {b_j})}
\end{equation}
where $f(.)$ is an activation function, ${\textbf{w}_j} = [{w_{j1}},{w_{j2}},...,{w_{jd}}] \in {\mathbb{R}^d}$ and ${b_j} \in \mathbb{R}$ ($i = 1,2,...,L$ ) are the learning parameters of the $i$th hidden node, ${\textbf{w}_j} \cdot {\textbf{x}_i}$ denotes the inner product of ${\textbf{w}_j}$ and ${\textbf{x}_i}$, and ${\beta_j} \in \mathbb{R}$ represents the weight parameter of the link connecting the $j$th hidden node to the output node. The above equation can be written in a vector form as
\begin{equation}
\textbf{Y} = \textbf{H}\boldsymbol{\beta}
\end{equation}
where $\textbf{Y} = {({y_1},...,{y_N})^T}$, $\boldsymbol{\beta}  = {({\beta _1},...,{\beta _L})^T}$ and
\begin{equation}
{\textbf{H}} = \left( \begin{array}{l}
f({\textbf{w}_1} \cdot {\textbf{x}_1} + {b_1}),\;\;\;...\;\;\\
\;\;\;\;\;\;\;\;\; \vdots \;\;\;\;\;\;\;\;\;\;\;\;\; \ddots \\
f({\textbf{w}_1} \cdot {\textbf{x}_N} + {b_1}),\;\;\;...\;
\end{array} \right.\left. \begin{array}{l}
f({\textbf{w}_L} \cdot {\textbf{x}_1} + {b_L})\\
\;\;\;\;\;\;\;\;\;\; \vdots \\
f({\textbf{w}_L} \cdot {\textbf{x}_N} + {b_L})
\end{array} \right)
\end{equation}
represents the output matrix of the hidden layer. In general, the output weight vector $\boldsymbol{\beta}$ can be solved by minimizing the regularized MSE (or least squares) loss:
\begin{equation}
{J_{MSE}}(\boldsymbol{\beta} ) = \sum\limits_{i = 1}^N {e_i^2}  + \lambda \left\| \boldsymbol{\beta}  \right\|_2^2 = \left\| {{\textbf{H}}\boldsymbol{\beta}  - \textbf{T}} \right\|_2^2 + \lambda \left\| \boldsymbol{\beta}  \right\|_2^2
\end{equation}
where ${e_i} = {t_i} - {y_i}$ is the error between the $i$th target response and the $i$th actual output, $\lambda \ge 0$ stands for the regularization parameter to prevent overfitting, and $\textbf{T} = {({t_1},...,{t_N})^T}$ is the target response vector. With a pseudo inversion operation, one can easily obtain a unique solution under the loss (20), that is
\begin{equation}
\boldsymbol{\beta}  = {[{{\textbf{H}}^\textbf{T}}{\textbf{H}} + \lambda {\textbf{I}}]^{ - 1}}{{\textbf{H}}^T}{\textbf{T}}
\end{equation}

\par In order to obtain a solution that is robust with respect to large outliers, now we consider the following KMPE based loss function:
\begin{equation}
\begin{aligned}
{J_{KMPE}}(\boldsymbol{\beta} ) &= {{\hat C}_p}(\textbf{T},{\textbf{H}}\boldsymbol{\beta} ) + \lambda \left\| \boldsymbol{\beta}  \right\|_2^2\\
&{\rm{              }} = \frac{1}{N}\sum\limits_{i = 1}^N {{{\left( {1 - {\kappa _\sigma }({e_i})} \right)}^{{p \mathord{\left/
 {\vphantom {p 2}} \right.
 \kern-\nulldelimiterspace} 2}}}}  + \lambda \left\| \boldsymbol{\beta}  \right\|_2^2\\
&{\rm{              }} = \frac{1}{N}\sum\limits_{i = 1}^N {{{\left( {1 - \exp ( - \frac{{e_i^2}}{{2{\sigma ^2}}})} \right)}^{{p \mathord{\left/
 {\vphantom {p 2}} \right.
 \kern-\nulldelimiterspace} 2}}}}  + \lambda \left\| \boldsymbol{\beta}  \right\|_2^2
\end{aligned}
\end{equation}
Note that different from the loss function in (20), the new loss function will be little influenced by large errors since the term ${\left( {1 - {\kappa _\sigma }({e_i})} \right)^{{p \mathord{\left/
 {\vphantom {p 2}} \right.
 \kern-\nulldelimiterspace} 2}}}$ is upper bounded by 1.0.

\par Let $\frac{\partial }{{\partial \boldsymbol{\beta} }}{J_{KMPE}}(\boldsymbol{\beta} ) = 0$. Then we derive

\begin{equation}
\begin{aligned}
&\frac{{\partial {J_{KMPE}}(\boldsymbol{\beta} )}}{{\partial \boldsymbol{\beta} }} = 0\\
&\Rightarrow \frac{1}{N}\sum\limits_{i = 1}^N {\left[ {\frac{{ - p}}{{2{\sigma ^2}}}{{\left( {1 - {\kappa _\sigma }({e_i})} \right)}^{({{p - 2)} \mathord{\left/
 {\vphantom {{p - 2)} 2}} \right.
 \kern-\nulldelimiterspace} 2}}}{\kappa _\sigma }({e_i}){e_i}\emph{\textbf{h}}_i^T} \right]}  + 2\lambda \boldsymbol{\beta}  = 0\\
& \Rightarrow \sum\limits_{i = 1}^N {\left[ { - {{\left( {1 - {\kappa _\sigma }({e_i})} \right)}^{({{p - 2)} \mathord{\left/
 {\vphantom {{p - 2)} 2}} \right.
 \kern-\nulldelimiterspace} 2}}}{\kappa _\sigma }({e_i}){e_i}\emph{\textbf{h}}_i^T} \right]}  + \frac{{4{\sigma ^2}N\lambda }}{p}\boldsymbol{\beta}  = 0\\
& \Rightarrow \sum\limits_{i = 1}^N {\left( {\varphi \left( {{e_i}} \right)\emph{\textbf{h}}_i^T{\emph{\textbf{h}}_i}{\boldsymbol{\beta}} - \varphi \left( {{e_i}} \right){t_i}\emph{\textbf{h}}_i^T} \right) + \lambda '\boldsymbol{\beta} }  = 0\\
& \Rightarrow \sum\limits_{i = 1}^N {\left( {\varphi \left( {{e_i}} \right){\emph{\textbf{h}}_i}^T{\emph{\textbf{h}}_i}\boldsymbol{\beta} } \right) + \lambda '\boldsymbol{\beta}  = \sum\limits_{i = 1}^N {\varphi \left( {{e_i}} \right){t_i}\emph{\textbf{h}}_i^T} } \\
& \Rightarrow \boldsymbol{\beta}  = {[{{\textbf{H}}^T}{\boldsymbol{\Lambda} \textbf{H}} + \lambda '{\textbf{I}}]^{ - 1}}{{\textbf{H}}^T}{\bf{\Lambda} \textbf{T}}
\end{aligned}
\end{equation}
where $\lambda ' = \frac{{4{\sigma ^2}N}}{p}\lambda $ is the $i$th row of \textbf{H}, $\varphi \left( {{e_i}} \right) = {\left( {1 - {\kappa _\sigma }({e_i})} \right)^{({{p - 2)} \mathord{\left/
 {\vphantom {{p - 2)} 2}} \right.
 \kern-\nulldelimiterspace} 2}}}{\kappa _\sigma }({e_i})$, and ${\bf{\Lambda }}$ is a diagonal matrix with diagonal elements ${{\bf{\Lambda }}_{ii}} = \varphi \left( {{e_i}} \right)$ .

\par The derived optimal solution $\boldsymbol{\beta}  = {[{{\textbf{H}}^T}{\bf{\Lambda}\textbf{ H}} + \lambda '{\textbf{I}}]^{ - 1}}{{\textbf{H}}^T}{\bf{\Lambda}\textbf{ T}}$ is not a closed-form solution since the matrix $\bf{\Lambda}$ on the right-hand side depends on the weight vector $\boldsymbol{\beta}$ through ${e_i} = {t_i} - {\emph{\textbf{h}}_i}\boldsymbol{\beta} $. So it is actually a fixed-point equation. The true optimal solution can thus be solved by a fixed-point iterative algorithm, as summarized in Algorithm 1. This algorithm is referred to as the ELM-KMPE in this work.

\begin{algorithm}[htb]
\scriptsize
\caption{ELM-KMPE}
\begin{algorithmic}[1]
\REQUIRE
samples $\{ {\textbf{x}_i},{t_i}\} _{i = 1}^N$
\ENSURE
weight vector $\boldsymbol{\beta}$\\
\noindent\textbf{Parameters setting}:number of hidden nodes $L$, regularization parameter $\lambda '$, maximum iteration number $M$, kernel width $\sigma$, power parameter $p$ and termination tolerance $\varepsilon $\\
\noindent\textbf{Initialization}: Set ${\boldsymbol{\beta} _\textbf{0}}{\rm{ = }}0$ and randomly initialize the parameters ${\textbf{w}_j}$ and ${{b}_j}$ ( $j = 1,...,L$)\\
\FOR {$k = 1,2,...,M$}          
    \STATE Compute the error based on ${\boldsymbol{\beta} _{k - 1}}$: ${e_i} = {t_i} - {\emph{\textbf{h}}_i}{\boldsymbol{\beta} _{k - 1}}$ \\
    \STATE Compute the diagonal matrix $\boldsymbol{\Lambda}$: ${{\boldsymbol{\Lambda }}_{ii}} = \varphi \left( {{e_i}} \right)$
    \STATE Update the weight vector${\boldsymbol{\beta}}$: ${\boldsymbol{\beta} _k} = {[{{\textbf{H}}^T}{\bf{\Lambda }\textbf{H}} + \lambda '{\textbf{I}}]^{ - 1}}{{\textbf{H}}^T}{\bf{\Lambda}\textbf{ T}}$\\
    \STATE \textbf{Until} $\left| {{J_{KMPE}}({\boldsymbol{\beta} _k}) - {J_{KMPE}}({\boldsymbol{\beta} _{k - 1}})} \right| < \varepsilon $
\ENDFOR

\end{algorithmic}
\end{algorithm}

\subsection{Principal Component Analysis}
The second example is the Principal Component Analysis (PCA), one of the most popular dimensionality reduction methods {\cite{jolliffe2002principal}}. Below we use the proposed KMPE as the loss function to derive a robust PCA algorithm.
\begin{spacing}{1.0}
\par Consider a set of samples ${\textbf{X}} = \left[ {{ \textbf{x} _1},...,{ \textbf{x} _n}} \right] \in {\mathbb{R}^{d \times n}}$, with $d$ being the dimension number and $n$ the sample number. The PCA methods try to find a projection matrix ${\textbf{W}} = [{\textbf{w}_1},...,{\textbf{w}_m}] \in {\mathbb{R}^{d \times m}}$ to define a new orthogonal coordinate system that can optimally describe the variability in the data set. In L2-PCA, the projection matrix is solved by minimizing the following loss function {\cite{jolliffe2002principal}}:\end{spacing}

\begin{equation}
\small
\begin{aligned}
{\ell _{L2}}\left( {\textbf{W}} \right) &= \left\| {{\bf{\tilde X}} - {\bf{WV}}} \right\|_2^2 = \sum\limits_{i = 1}^n {\left\| {{\textbf{x}_i} - \boldsymbol{\mu}  - \sum\limits_{k = 1}^m {{\textbf{w}_k}{v_{ki}}} } \right\|_2^2}  \\
&= \sum\limits_{i = 1}^n {\sum\limits_{j = 1}^d {{{\left( {{x_{ji}} - {\mu _j} - \sum\limits_{k = 1}^m {{w_{jk}}{v_{ki}}} } \right)}^2}} }
\end{aligned}
\end{equation}

\noindent where ${\bf{\tilde X}} = \left[ {{\bf{\tilde x}_1},...,{\bf{\tilde x}_n}} \right]$ denotes the column-wise-zero-mean version of $\textbf{X}$, with ${\bf{\tilde x}_i} = {\bf{x}_i} - \boldsymbol{\mu} $, $\boldsymbol{\mu} $ is the sample mean of column vectors, and ${\textbf{V}} = {{\textbf{W}}^T}{\bf{\tilde X}} = [{\textbf{v}_1},...,{\textbf{v}_n}] \in {\mathbb{R}^{m \times n}}$ contains the principal components that are projected under the projection matrix $\textbf{W}$.

\par In order to prevent the outliers in the edge data from corrupting the results of dimensionality reduction, we minimize the following robust cost function for PCA:
\begin{equation}
\begin{aligned}
&{\ell _{KMPE}}\left( {{\textbf{W}},\boldsymbol{\mu} } \right) = {{\hat C}_p}\left( {{\bf{\tilde X}},{\textbf{W}}{{\textbf{W}}^T}{\bf{\tilde X}}} \right)\\
&= \frac{1}{n}\sum\limits_{i = 1}^n {{{\left( {1 - {\kappa _\sigma }\left( {{\textbf{x}_i} -\boldsymbol{\mu}  - {\textbf{W}}{{\textbf{W}}^T}\left( {{\textbf{x}_i} - \boldsymbol{\mu} } \right)} \right)} \right)}^{{p \mathord{\left/
 {\vphantom {p 2}} \right.
 \kern-\nulldelimiterspace} 2}}}} \\
&= \frac{1}{n}\sum\limits_{i = 1}^n {{{\left( {1 - \exp \left( { - \frac{{\left\| {{\textbf{e}_i}} \right\|_2^2}}{{2{\sigma ^2}}}} \right)} \right)}^{{p \mathord{\left/
 {\vphantom {p 2}} \right.
 \kern-\nulldelimiterspace} 2}}}} \\
&= \frac{1}{n}\sum\limits_{i = 1}^n {\rho \left( {{{\left\| {{\textbf{e}_i}} \right\|}_2}} \right)}
\end{aligned}
\end{equation}
where ${\textbf{e}_i} = {\textbf{x}_i} - \boldsymbol{\mu}  - {\textbf{W}}{{\textbf{W}}^T}\left( {{\textbf{x}_i} - \boldsymbol{\mu} } \right)$. Indeed, the cost function $\rho \left( {{{\left\| {{\textbf{e}_i}} \right\|}_2}} \right) = {\left( {1 - \exp \left( { - \frac{{\left\| {{\textbf{e}_i}} \right\|_2^2}}{{2{\sigma ^2}}}} \right)} \right)^{{p \mathord{\left/
 {\vphantom {p 2}} \right.
 \kern-\nulldelimiterspace} 2}}}$ belongs to the M-estimation robust cost functions {\cite{maronna2006robust,huber1981wiley}}, and minimizing the cost (25) is an M-estimation problem. It is instructive and useful to transform the minimization of (25) into a weighted least squares problem, which can be solved by iteratively reweighted least squares (IRLS). This method is originally proposed in {\cite{beaton1974fitting}} and successfully used in robust statistics {\cite{holland1977robust}}, computer vision {\cite{lai2000robust,zhang1997parameter}}, face recognition {\cite{wei2015undersampled,he2011maximum}} and PCA {\cite{de2003framework}}. Here, the weighting matrix ${\bf{\Lambda }}$ is a diagonal matrix with elements ${\Lambda _{ii}} = {{\psi \left( {{{\left\| {{\textbf{e}_i}} \right\|}_2}} \right)} \mathord{\left/
 {\vphantom {{\psi \left( {{{\left\| {{\textbf{e}_i}} \right\|}_2}} \right)} {{{\left\| {{\textbf{e}_i}} \right\|}_2}}}} \right.
 \kern-\nulldelimiterspace} {{{\left\| {{\textbf{e}_i}} \right\|}_2}}}$, where $\psi \left( {{{\left\| {{\textbf{e}_i}} \right\|}_2}} \right) = \frac{{\partial \rho \left( {{{\left\| {{\textbf{e}_i}} \right\|}_2}} \right)}}{{\partial {{\left\| {{\textbf{e}_i}} \right\|}_2}}}$. In this way, the cost function (25) will be equivalent to the following weighted least squares cost:
\begin{equation}
\begin{aligned}
&{{\tilde \ell }_{KMPE}}\left( {\textbf{W},\boldsymbol{\mu} } \right)\\
&=\! \sum\limits_{i = 1}^n {{{\!\left( {{\textbf{x}_i} \!-\! \boldsymbol{\mu}  \!-\! {\textbf{W}}{\textbf{W}^T}\left( {{\textbf{x}_i} \!-\! \boldsymbol{\mu} } \right)} \right)}^T}{\Lambda _{ii}}\left( {{\textbf{x}_i} \!-\! \boldsymbol{\mu}  \!-\! {\textbf{W}}{\textbf{W}^T}\left( {{\textbf{x}_i} \!-\! \boldsymbol{\mu} } \right)}\! \right)}
\end{aligned}
\end{equation}

where
\begin{equation}
{\Lambda _{ii}} = {\left( {1 - \exp \left( { - \frac{{\left\| {{\textbf{e}_i}} \right\|_2^2}}{{2{\sigma ^2}}}} \right)} \right)^{{{(p - 2)} \mathord{\left/
 {\vphantom {{(p - 2)} 2}} \right.
 \kern-\nulldelimiterspace} 2}}}\exp \left( { - \frac{{\left\| {{\textbf{e}_i}} \right\|_2^2}}{{2{\sigma ^2}}}} \right)
\end{equation}
Setting $\frac{\partial }{{\partial \boldsymbol{\mu} }}{\tilde \ell _{KMPE}}\left( {\textbf{W},\boldsymbol{\mu} } \right){\rm{ = }}\textbf{0}$, we derive
\begin{equation}
\boldsymbol{\mu}  = {{\sum\limits_{i = 1}^n {{\Lambda _{ii}}{\textbf{x}_i}} } \mathord{\left/
 {\vphantom {{\sum\limits_{i = 1}^n {{\Lambda _{ii}}{\textbf{x}_i}} } {\sum\limits_{i = 1}^n {{\Lambda _{ii}}} }}} \right.
 \kern-\nulldelimiterspace} {\sum\limits_{i = 1}^n {{\Lambda _{ii}}} }}
\end{equation}
In addition, we can easily obtain the following solution
\begin{equation}
{\textbf{W}} = \mathop {\arg \max }\limits_{\textbf{W}} Tr\left( {{{\textbf{W}}^T}{\bf{\tilde X\Lambda }}{{{\bf{\tilde X}}}^T}{\textbf{W}}} \right)
\end{equation}
The optimization problem (29) is a weighted PCA that can be computed by solving the corresponding eigenvalue problem. The solution of (25) can thus be obtained by iterating (27), (28) and (29). This algorithm is called in this work the PCA-KMPE, which when $p=2.0$ will perform the HQ-PCA {\cite{he2011robust}}. To learn an $m$-dimensional subspace, one can use a trick as in {\cite{he2011robust}} to learn a small $m_r$ dimensional subspace to further eliminate the influence by outliers. The proposed PCA-KMPE is summarized in Algorithm 2.

\begin{algorithm}[htb]
\scriptsize
\caption{PCA-KMPE}
\begin{algorithmic}[1]
\REQUIRE
input data $\textbf{X}$
\ENSURE
projection matrix ${\textbf{W}} \in {R^{d \times m}}$\\
\noindent\textbf{Parameters setting}: maximum iteration number $M$, kernel width $\sigma$, power parameter $p$ and termination tolerance $\varepsilon $\\
\noindent\textbf{Initialization}: ${\boldsymbol{\mu} _0} = \frac{1}{n}\sum\limits_{i = 1}^n {{\textbf{x}_i}} $, ${\textbf{W}_0} = {\textbf{W}_{PCA}}$(solution of the original PCA)\\
\FOR {$k = 1,2,...,M$}          
    \STATE Compute the errors based on $\textbf{W}_{k-1}$ and ${\boldsymbol{\mu} _{k-1}}$: \\
    ${\textbf{e}_i} = \left( {{\textbf{x}_i} - {\boldsymbol{\mu} _{\;k - 1}}} \right) - {{\textbf{W}}_{k - 1}}{\textbf{W}}_{k - 1}^T\left( {{\textbf{x}_i} - {\boldsymbol{\mu} _{\;k - 1}}} \right)$\\
    \STATE Compute the diagonal matrix $\boldsymbol{\Lambda}$ using (27)
    \STATE Update the sample mean using (28)
    \STATE Update the projection matrix $\textbf{w}_k$ by solving the eigenvalue problem (29)
    \STATE \textbf{Until} ${\left\| {{{\textbf{W}}_{k - 1}} - {{\textbf{W}}_k}} \right\|_2} \le \varepsilon $
\ENDFOR

\end{algorithmic}
\end{algorithm}
The kernel width $\sigma$ is an important parameter in PCA-KMPE. In general, one can employ the Silverman’s rule {\cite{silverman1986density}}, to adjust the kernel width:
\begin{equation}
{\sigma ^2} = 1.06 \times \min \left\{ {{\sigma _E},\frac{R}{{1.354}}} \right\} \times {(n)^{{{ - 1} \mathord{\left/
 {\vphantom {{ - 1} 5}} \right.
 \kern-\nulldelimiterspace} 5}}}
\end{equation}
where $\sigma_E$ is the standard deviation of $\left\| {{\textbf{e}_i}} \right\|_2^2$ and $R$ is the interquartile range.

\section{EXPERIMENTAL  RESULTS}
This section presents some experimental results to verify the advantages of the ELM-KMPE and PCA-KMPE developed in the previous section.
\subsection{Function estimation with synthetic data}
In this example, the sinc function estimation, a popular illustration example for nonlinear regression problem in the literature, is used to evaluate the performance of the proposed ELM-KMPE and other ELM algorithms, such as ELM [21], RELM {\cite{huang2012extreme}} and ELM-RCC {\cite{xing2013training}}. The synthetic data are generated by $y(i) = k \cdot {\mathop{\rm sinc}\nolimits} \left( {x(i)} \right) + v(i)$, where $k=8$,
\begin{equation}
sinc(x) = \left\{ \begin{array}{l}
\sin (x)/x\;\;\;x \ne 0\\
1\;\;\;\;\;\;\;\;\;\;\;\;\;\;x = 0
\end{array} \right.
\end{equation}
and $v(i)$ is a noise modeled as $v(i) = (1 - a(i))A(i) + a(i)B(i)$, where $a(i)$ is a binary iid process with probability mass $\Pr \left\{ {a(i) = 1} \right\} = c$, $\Pr \left\{ {a(i) = 0} \right\} = 1 - c$ ($0 \le c \le 1$ ), $A(i)$ denotes the background noise and $B(i)$ is another noise process to represent outliers. The noise processes $A(i)$ and $B(i)$ are mutually independent and both independent of $a(i)$. In this subsection, $c$ is set at 0.1 and $B(i)$ is assumed to be a zero-mean Gaussian noise with variance 9.0. Two background noises are considered: a) Uniform distribution over $[-1.0, 1.0]$ and b) Sine wave noise $\sin \left( \omega  \right)$, with $\omega$ uniformly distributed over $\left[ {0,2\pi } \right]$. In addition, the input data $x(i)$ are drawn uniformly from $[ - 10,10]$. In the simulation, 200 samples are used for training and another 200 noise-free samples are used for testing. The RMSE is employed to measure the performance, calculated by
\begin{equation}
RMSE = \sqrt {\frac{1}{N}\sum\limits_{i = 1}^N {{{({y_i} - {{\tilde y}_i})}^2}} }
\end{equation}
where ${y_i}$ and ${\tilde y_i}$ denote the target values and corresponding estimated values respectively, and $N$ is the number of samples. The parameter settings of four algorithms under two distributions of $A(i)$ are summarized in Table 1, where $L$, $\lambda$ (or $\lambda '$), $\sigma$ and $p$ denote the number of hidden layer nodes, regularization parameter, kernel width and the power parameter in ELM-KMPE. The estimation results and testing RMSEs are illustrated in Fig.1 and Table.2. It is evident that the ELM-KMPE achieves the best performance among the four algorithms.
\begin{figure*}[htbp]
\setlength{\abovecaptionskip}{0pt}
\setlength{\belowcaptionskip}{0pt}
\centering
\subfigure[]{
\includegraphics[width=3.0in,height=2.4in]{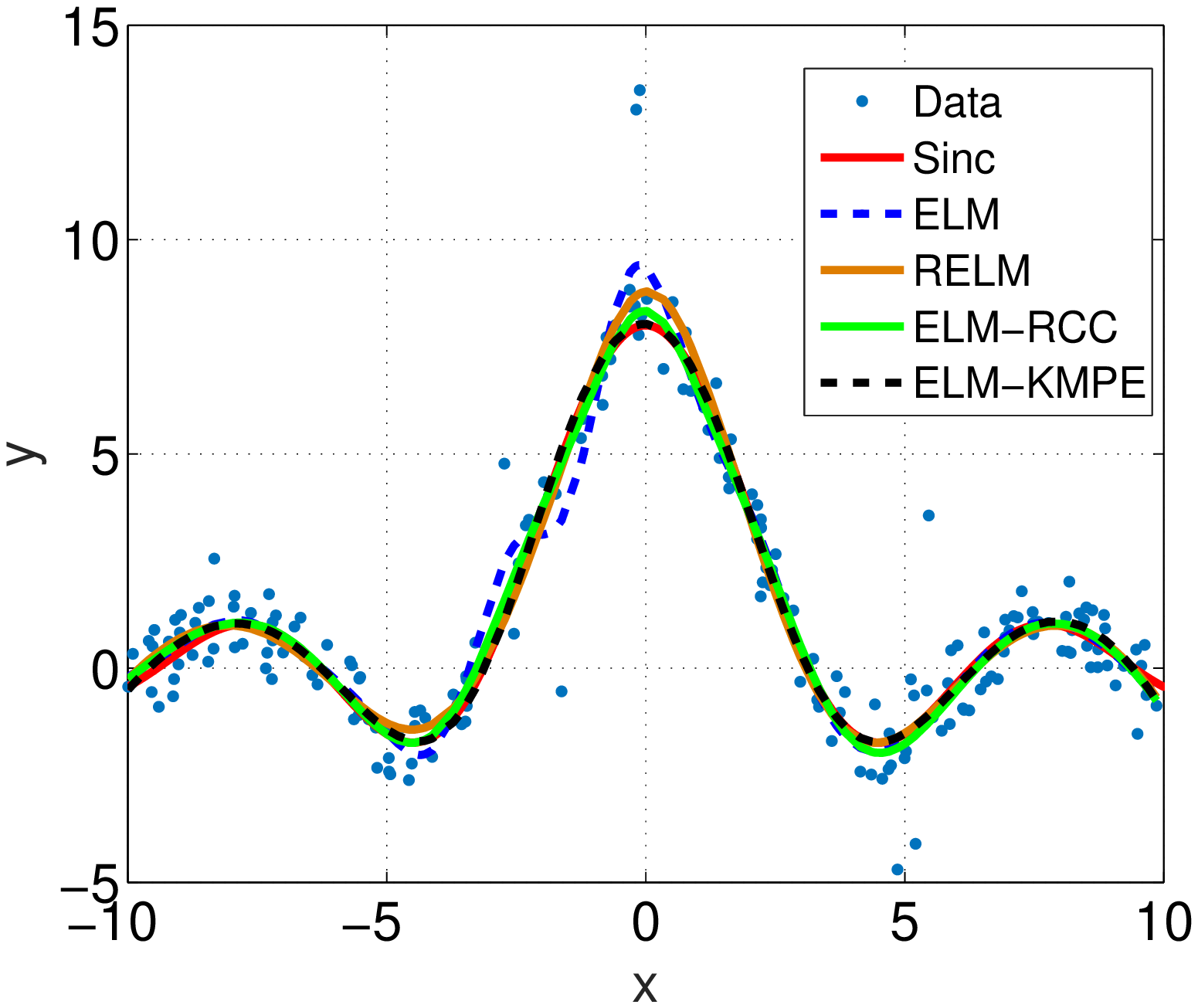}}
\subfigure[]{
\includegraphics[width=3.0in,height=2.4in]{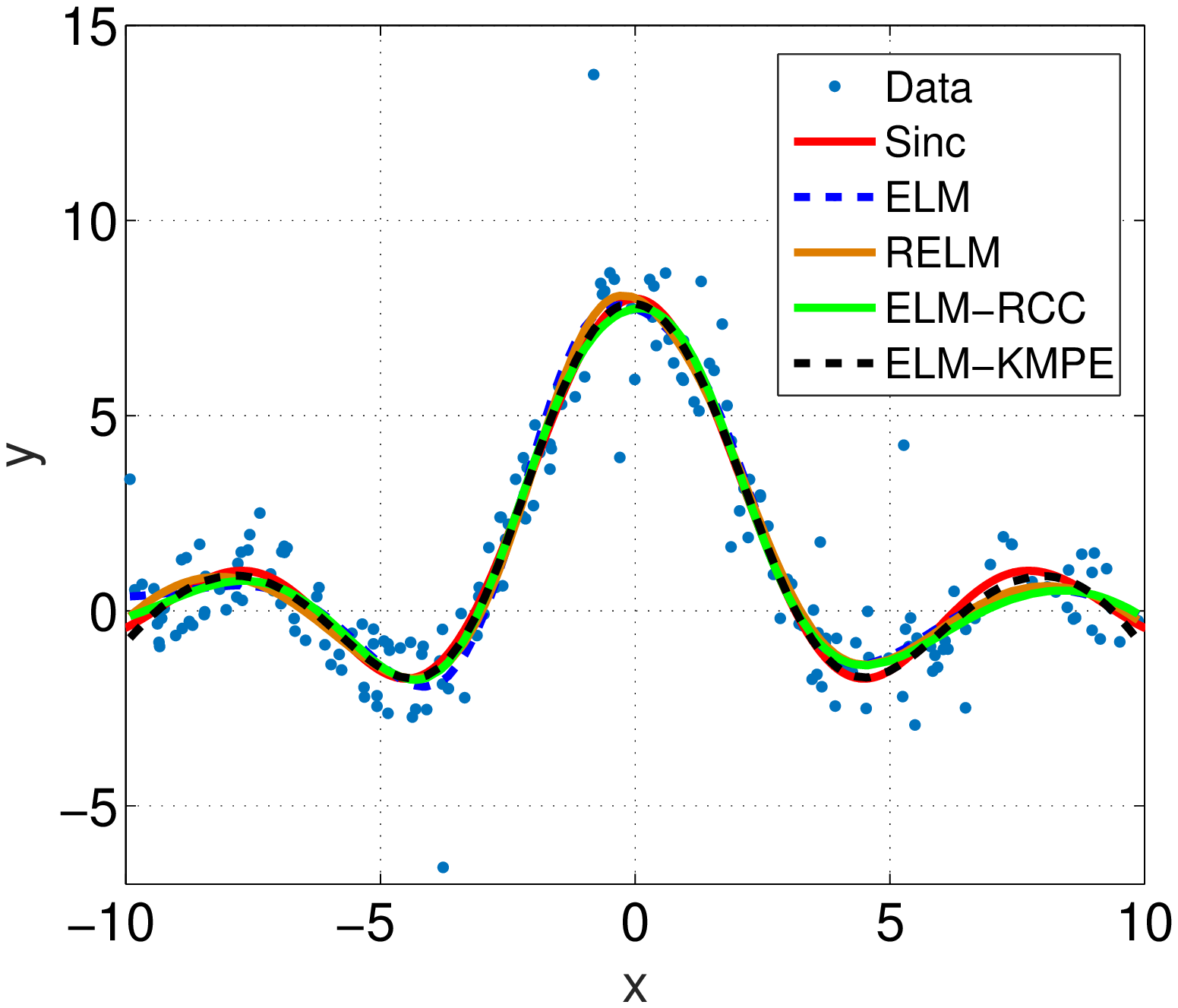}}
\caption{Sinc function estimation results with different background noises: (a) Uniform (b) Sine wave}
\label{fig1}
\end{figure*}

\begin{table*}[]\small
\renewcommand\arraystretch{1.5}
\setlength{\abovecaptionskip}{0pt}
\setlength{\belowcaptionskip}{5pt}
\centering
\caption{Parameter settings of four algorithms under two distributions of $A(i)$}
\begin{tabular}{cccccccccccc}
\toprule
 &$\quad$&ELM&\multicolumn{2}{c}{RELM} & \multicolumn{3}{c}{ELM-RCC} & \multicolumn{4}{c}{ELM-KMPE}\\
\hline
 &$\quad$&$L$&$L$&$\lambda$&$L$&$\lambda$&$\sigma$&$L$&$\lambda '$&$\sigma$&$p$\\
\hline
&Uniform&20&90&$5 \times {10^{ - 5}}$&90&${10^{ - 6}}$&1.5&90&$2 \times {10^{ - 6}}$&0.8&4\\
&Sine wave&10&40&$5 \times {10^{ - 5}}$&25&$5 \times {10^{ - 6}}$&2&25&$2.5 \times {10^{ - 6}}$&1.2&3.4\\
\bottomrule
\end{tabular}
\end{table*}

\begin{table}[]\small
\renewcommand\arraystretch{1.5}
\setlength{\abovecaptionskip}{0pt}
\setlength{\belowcaptionskip}{5pt}
\centering
\caption{Testing RMSEs of four algorithms}
\begin{tabular}{cccccc}
\toprule
&$\quad$& ELM & RELM & ELM-RCC &ELM-KMPE\\
\hline
&Uniform &	0.5117&	0.2234&	0.1671&	\textbf{0.1079}\\
&Sine wave&	0.3340&	0.2498&	0.2335&	\textbf{0.1156}\\
\bottomrule
\end{tabular}
\end{table}

\subsection{Regression and classification on benchmark datasets}
In this subsection, we compare the aforementioned four algorithms in regression and classification problems with benchmark datasets from UCI machine learning repository {\cite{frank2010uci}}. The details of the datasets are shown in Table 3 and 4. For each dataset, the training and testing samples are randomly selected form the set. In particular, the data for regression are normalized to the range $[0, 1]$. The parameter settings of the four algorithms for regression and classification experiments are presented in Table 5 and 6. For each algorithm, the parameters are experimentally chosen by fivefold cross-validation. The RMSE is used as the performance measure for regression. For classification, the performance is measured by the accuracy (ACC). Let $p_i$ and $t_i$ be the predicted and target labels of the $i$th sample. The ACC is defined by
\begin{equation}
ACC = \frac{1}{n}\sum\limits_{i = 1}^n {\delta ({t_i},map({p_i}))}
\end{equation}
where $\delta (x,y)$ is an indicator function, $\delta (x,y)=1$ if $x=y$, otherwise $\delta (x,y)=0$, and $map( \cdot )$ maps each predicted label to the equivalent target label. The Kuhn-Munkres algorithm {\cite{lovasz2009matching}} is employed to realize such a mapping. The ``mean $ \pm $ standard deviation'' results of the RMSE and ACC during training and testing are shown in Table 7 and 8, where the best testing results are represented in bold for each data set. As one can see, in all the cases the proposed ELM-KMPE can outperform other algorithms.

\begin{table}[]\small
\renewcommand\arraystretch{1.5}
\setlength{\abovecaptionskip}{0pt}
\setlength{\belowcaptionskip}{0pt}
\centering
\caption{Specification of the regression problem}
\begin{tabular}{cccc}
\toprule
\multirow{2}*{Datasets} & \multirow{2}*{Features} & \multicolumn{2}{c}{Observations}\\
\cline{3-4}
&{}&Training&Testing\\
\midrule
Servo	&5	&83	&83\\
Concrete	&9	&515	&515\\
Wine red	&12	&799	&799\\
Housing	&14	&253	&253\\
Airfoil	&5	&751	&751\\
Slump	&10	&52	&51\\
Yacht	&6	&154	&154\\
\bottomrule
\end{tabular}
\end{table}

\begin{table}[]\small
\renewcommand\arraystretch{1.5}
\setlength{\abovecaptionskip}{0pt}
\setlength{\belowcaptionskip}{5pt}
\centering
\caption{Specification of the classification problem}
\begin{tabular}{ccccc}
\toprule
\multirow{2}*{Datasets} & \multirow{2}*{Classes} & \multirow{2}*{Features} & \multicolumn{2}{c}{Observations}\\
\cline{4-5}
&{}&{}&Training&Testing\\
\midrule
Glass	&7	&11	&114	&100\\
Wine	&3	&13	&89	&89\\
Ecoli	&8	&7	&180	&156\\
User-Modeling	&2	&5	&138	&120\\
Wdbc	&2	&30	&100	&496\\
Leaf	&36	&14	&180	&160\\
Vehicle	&4	&18	&500	&346\\
Seed	&3	&7	&110	&100\\
\bottomrule
\end{tabular}
\end{table}

\begin{table*}[]\small
\renewcommand\arraystretch{1.5}
\setlength{\abovecaptionskip}{0pt}
\setlength{\belowcaptionskip}{5pt}
\centering
\caption{Parameter settings of four algorithms in regression}
\begin{tabular}{ccccccccccc}
\toprule
\multirow{2}*{Datasets} & ELM & \multicolumn{2}{c}{RELM}& \multicolumn{3}{c}{ELM-RCC}& \multicolumn{4}{c}{ELM-KMPE}\\
\cline{2-11}
&L&L&$\lambda$&L&$\lambda$&$\sigma$&L&$\lambda '$&$\sigma$&p\\
\hline
Servo	&25&90&0.00001&65&0.8&0.0001&75&0.9&0.00001&1.6\\
Concrete	&120&185&0.0002&200&0.6&0.000005&200&0.7&0.00005&2.2\\
Wine red	&15	&15&0.000002&25&0.3&0.001&115&0.5&0.001&2.2\\
Housing	&40	&180&0.001&200&0.8&0.001&200&0.9&0.002&2.2\\
Airfoil	&130&200&0.0002	&150	&0.4	&0.0000001	&195	&1.2	&0.0000001	&2.4\\
Slump	&195&190&0.000025&165	&0.6	&0.000001	&190	&0.4	&0.000002	&2.8\\
Yacht	&90	&185&0.000025&195	&0.4	&0.0000001	&175	&1	&0.0000001&	1.0\\
\bottomrule
\end{tabular}
\end{table*}

\begin{table*}[]\small
\renewcommand\arraystretch{1.5}
\setlength{\abovecaptionskip}{0pt}
\setlength{\belowcaptionskip}{5pt}
\centering
\caption{Parameter settings of four algorithms in classification}
\begin{tabular}{ccccccccccc}
\toprule
\multirow{2}*{Datasets} & ELM & \multicolumn{2}{c}{RELM}& \multicolumn{3}{c}{ELM-RCC}& \multicolumn{4}{c}{ELM-KMPE}\\
\cline{2-11}
&L&L&$\lambda$&L&$\lambda$&$\sigma$&L&$\lambda '$&$\sigma$&p\\
\hline
Glass	&105	&195	&0.00005	&185	&1.5	&0.001	&180	&1.4	&0.001	&2.8\\
Wine	&15	&15	&0.000025	&20	&1.4	&0.001	&25	&0.8	&0.001	&2.8\\
Ecoli	&10	&90&	0.001	&155	&1.8	&0.001	&25	&1.4	&0.00005	&2.8\\
User-
Modeling	&40	&145	&0.000025	&125	&1.8	&0.000005	&70	&0.9	&0.000002	&2.4\\
Wdbc	&40	&145	&0.000025	&125	&1.5	&0.00005	&50	&1.2	&0.00005	&2.2\\
Leaf	&70	&130	&0.00001	&200	&1.7	&0.000025	&180	&1.3	&0.0001	&2.8\\
Vehicle	&130	&155	&0.00001	&195	&1.3	&0.00001	&200	&1	&0.00005	&2.4\\
Seed	&30	&130	&0.0001&	200	&1.5	&0.001	&170	&1.5	&0.001	&2.8\\
\bottomrule
\end{tabular}
\end{table*}

\begin{table*}[]\footnotesize
\renewcommand\arraystretch{1.5}
\setlength{\abovecaptionskip}{0pt}
\setlength{\belowcaptionskip}{5pt}
\centering
\caption{Performance comparison of four algorithms with benchmark regression datasets}
\begin{tabular}{p{1cm}p{1cm}p{1cm}p{1cm}p{1cm}p{1cm}p{1cm}p{1cm}p{1cm}}
\toprule
\multicolumn{1}{c}{\multirow{3}*{Datasets}} & \multicolumn{2}{c}{ELM} & \multicolumn{2}{c}{RELM}& \multicolumn{2}{c}{ELM-RCC}& \multicolumn{2}{c}{ELM-KMPE}\\
\cline{2-9}
&\multicolumn{1}{c}{Training} &\multicolumn{1}{c}{Testing} &\multicolumn{1}{c}{Training} &\multicolumn{1}{c}{Testing}&\multicolumn{1}{c}{Training} &\multicolumn{1}{c}{Testing}&\multicolumn{1}{c}{Training} &\multicolumn{1}{c}{Testing} \\
&\multicolumn{1}{c}{RMSE}&\multicolumn{1}{c}{RMSE}&\multicolumn{1}{c}{RMSE}&\multicolumn{1}{c}{RMSE}&\multicolumn{1}{c}{RMSE}&\multicolumn{1}{c}{RMSE}&\multicolumn{1}{c}{RMSE}&\multicolumn{1}{c}{RMSE}\\
\hline
\multicolumn{1}{c}{Servo}	&\multicolumn{1}{c}{0.0741$\pm$0.0126}	&\multicolumn{1}{c}{0.1183$\pm$0.0204}	&\multicolumn{1}{c}{0.0720$\pm$0.0106}	&\multicolumn{1}{c}{0.1036$\pm$0.0152}	&\multicolumn{1}{c}{0.0739$\pm$0.0106}	&\multicolumn{1}{c}{0.1032$\pm$0.0148}	&\multicolumn{1}{c}{0.0570$\pm$0.0108}	&\multicolumn{1}{c}{\textbf{0.1022$\pm$0.0184}}\\
\multicolumn{1}{c}{Concrete}	&\multicolumn{1}{c}{0.0612$\pm$0.0026}	&\multicolumn{1}{c}{0.0994$\pm$0.0013}	&\multicolumn{1}{c}{0.0742$\pm$0.0025}	&\multicolumn{1}{c}{0.0914$\pm$0.0042}	&\multicolumn{1}{c}{0.0559$\pm$0.0018}	&\multicolumn{1}{c}{0.0879$\pm$0.0077}	&\multicolumn{1}{c}{0.0577$\pm$0.0021}	&\multicolumn{1}{c}{\textbf{0.0864$\pm$0.0058}}\\
\multicolumn{1}{c}{Wine red}	&\multicolumn{1}{c}{0.1280$\pm$0.0031}	&\multicolumn{1}{c}{0.1312$\pm$0.0032}	&\multicolumn{1}{c}{0.1282$\pm$0.0031}	&\multicolumn{1}{c}{0.1309$\pm$0.0031}	&\multicolumn{1}{c}{0.1264$\pm$0.0031}	&\multicolumn{1}{c}{0.1306$\pm$0.0032}	&\multicolumn{1}{c}{0.1198$\pm$0.0028}	&\multicolumn{1}{c}{\textbf{0.1302$\pm$0.0035}}\\
\multicolumn{1}{c}{Housing}&\multicolumn{1}{c}{0.0728$\pm$0.0070}	&\multicolumn{1}{c}{0.0994$\pm$0.0120}	&\multicolumn{1}{c}{0.0502$\pm$0.0044}	&\multicolumn{1}{c}{0.0835$\pm$0.0100}	&\multicolumn{1}{c}{0.0493$\pm$0.0046}	&\multicolumn{1}{c}{0.0832$\pm$0.0099}	&\multicolumn{1}{c}{0.0554$\pm$0.0045}	&\multicolumn{1}{c}{\textbf{0.0821$\pm$0.0101}}\\
\multicolumn{1}{c}{Airfoil}&\multicolumn{1}{c}{0.0664$\pm$0.0027}	&\multicolumn{1}{c}{0.0942$\pm$0.0099}	&\multicolumn{1}{c}{0.0967$\pm$0.0061}	&\multicolumn{1}{c}{0.1025$\pm$0.0058}	&\multicolumn{1}{c}{0.0742$\pm$0.0026}	&\multicolumn{1}{c}{0.0896$\pm$0.0050}	&\multicolumn{1}{c}{0.0695$\pm$0.0029}	&\multicolumn{1}{c}{\textbf{0.0880$\pm$0.0058}}\\
\multicolumn{1}{c}{Slump}	&\multicolumn{1}{c}{0$\pm$0}	&\multicolumn{1}{c}{0.0429$\pm$0.0091}	&\multicolumn{1}{c}{0.0066$\pm$0.0041}	&\multicolumn{1}{c}{0.0424$\pm$0.0097}	&\multicolumn{1}{c}{0.0001$\pm$0}	&\multicolumn{1}{c}{0.0423$\pm$0.0130}	&\multicolumn{1}{c}{0.0028$\pm$0.0004}	&\multicolumn{1}{c}{\textbf{0.0410$\pm$0.0107}}\\
\multicolumn{1}{c}{Yacht}	&\multicolumn{1}{c}{0.0040$\pm$0.0004}	&\multicolumn{1}{c}{0.0740$\pm$0.1267}	&\multicolumn{1}{c}{0.0370$\pm$0.0079}	&\multicolumn{1}{c}{0.0530$\pm$0.0086}	&\multicolumn{1}{c}{0.0126$\pm$0.0008}	&\multicolumn{1}{c}{0.0333$\pm$0.0086}	&\multicolumn{1}{c}{0.0051$\pm$0.0009}	&\multicolumn{1}{c}{\textbf{0.0250$\pm$0.0147}}\\

\bottomrule
\end{tabular}
\end{table*}

\begin{table*}[]\small
\renewcommand\arraystretch{1.5}
\setlength{\abovecaptionskip}{0pt}
\setlength{\belowcaptionskip}{5pt}
\centering
\caption{Performance comparison of four algorithms with benchmark classification datasets}
\begin{tabular}{p{1cm}p{1.6cm}p{1.6cm}p{1.6cm}p{1.6cm}p{1.6cm}p{1.6cm}p{1.6cm}p{1.6cm}}
\toprule
\multicolumn{1}{c}{\multirow{3}*{Datasets}} & \multicolumn{2}{c}{ELM} & \multicolumn{2}{c}{RELM}& \multicolumn{2}{c}{ELM-RCC}& \multicolumn{2}{c}{ELM-KMPE}\\
\cline{2-9}
&\multicolumn{1}{c}{Training} &\multicolumn{1}{c}{Testing} &\multicolumn{1}{c}{Training} &\multicolumn{1}{c}{Testing}&\multicolumn{1}{c}{Training} &\multicolumn{1}{c}{Testing}&\multicolumn{1}{c}{Training} &\multicolumn{1}{c}{Testing} \\
&\multicolumn{1}{c}{ACC}&\multicolumn{1}{c}{ACC}&\multicolumn{1}{c}{ACC}&\multicolumn{1}{c}{ACC}&\multicolumn{1}{c}{ACC}&\multicolumn{1}{c}{ACC}&\multicolumn{1}{c}{ACC}&\multicolumn{1}{c}{ACC}\\
\hline
\multicolumn{1}{c}{Glass}	&\multicolumn{1}{c}{95.32$\pm$3.39}	&\multicolumn{1}{c}{77.62$\pm$9.55}	&\multicolumn{1}{c}{96.04$\pm$1.72}	&\multicolumn{1}{c}{92.50$\pm$4.31}	&\multicolumn{1}{c}{96.80$\pm$2.71}	&\multicolumn{1}{c}{93.24$\pm$3.38}	&\multicolumn{1}{c}{97.85$\pm$2.03}	&\multicolumn{1}{c}{\textbf{94.54$\pm$3.12}}\\
\multicolumn{1}{c}{Wine}	&\multicolumn{1}{c}{99.55$\pm$0.70}	&\multicolumn{1}{c}{96.91$\pm$2.07}	&\multicolumn{1}{c}{99.65$\pm$0.67}	&\multicolumn{1}{c}{96.92$\pm$2.07}	&\multicolumn{1}{c}{99.85$\pm$0.44}	&\multicolumn{1}{c}{97.43$\pm$1.95}	&\multicolumn{1}{c}{99.91$\pm$0.35}	&\multicolumn{1}{c}{\textbf{97.58$\pm$1.73}}\\
\multicolumn{1}{c}{Ecoli}	&\multicolumn{1}{c}{90.45$\pm$3.46}	&\multicolumn{1}{c}{80.65$\pm$3.51}	&\multicolumn{1}{c}{92.61$\pm$3.44}	&\multicolumn{1}{c}{82.19$\pm$3.11}	&\multicolumn{1}{c}{92.48$\pm$3.24}	&\multicolumn{1}{c}{82.27$\pm$2.93}	&\multicolumn{1}{c}{93.50$\pm$3.39}	&\multicolumn{1}{c}{\textbf{82.35$\pm$2.77}}\\
\multicolumn{1}{c}{User-Modeling}	&\multicolumn{1}{c}{92.80$\pm$1.99}	&\multicolumn{1}{c}{84.17$\pm$3.63}	&\multicolumn{1}{c}{93.47$\pm$1.68}	&\multicolumn{1}{c}{85.47$\pm$3.26}	&\multicolumn{1}{c}{94.02$\pm$1.63}	&\multicolumn{1}{c}{85.57$\pm$3.42}	&\multicolumn{1}{c}{93.17$\pm$1.82}	&\multicolumn{1}{c}{\textbf{86.29$\pm$3.08}}\\
\multicolumn{1}{c}{Wdbc}	&\multicolumn{1}{c}{93.07$\pm$2.11}	&\multicolumn{1}{c}{84.81$\pm$3.43}	&\multicolumn{1}{c}{93.52$\pm$2.07}	&\multicolumn{1}{c}{85.86$\pm$3.31}	&\multicolumn{1}{c}{92.34$\pm$2.16}	&\multicolumn{1}{c}{86.63$\pm$3.27}	&\multicolumn{1}{c}{91.01$\pm$2.02}	&\multicolumn{1}{c}{\textbf{87.09$\pm$3.20}}\\
\multicolumn{1}{c}{Leaf}	&\multicolumn{1}{c}{93.63$\pm$1.28}	&\multicolumn{1}{c}{68.86$\pm$3.92}	&\multicolumn{1}{c}{95.91$\pm$1.15}	&\multicolumn{1}{c}{71.51$\pm$3.71}	&\multicolumn{1}{c}{95.01$\pm$1.53}	&\multicolumn{1}{c}{71.56$\pm$3.88}	&\multicolumn{1}{c}{95.21$\pm$1.48}	&\multicolumn{1}{c}{\textbf{73.87$\pm$4.20}}\\
\multicolumn{1}{c}{Vehicle}	&\multicolumn{1}{c}{92.87$\pm$1.03}	&\multicolumn{1}{c}{81.14$\pm$1.91}	&\multicolumn{1}{c}{94.01$\pm$0.97}	&\multicolumn{1}{c}{81.51$\pm$1.99}	&\multicolumn{1}{c}{94.88$\pm$0.80}	&\multicolumn{1}{c}{81.61$\pm$2.03}	&\multicolumn{1}{c}{95.80$\pm$0.79}	&\multicolumn{1}{c}{\textbf{82.23$\pm$2.25}}\\
\multicolumn{1}{c}{Seed}	&\multicolumn{1}{c}{98.48$\pm$1.07}	&\multicolumn{1}{c}{92.26$\pm$2.65}	&\multicolumn{1}{c}{98.75$\pm$0.93}	&\multicolumn{1}{c}{94.40$\pm$2.13}	&\multicolumn{1}{c}{98.30$\pm$1.03}	&\multicolumn{1}{c}{94.65$\pm$1.95}	&\multicolumn{1}{c}{98.65$\pm$1.05} &\multicolumn{1}{c}{\textbf{95.01$\pm$1.96}}\\

\bottomrule
\end{tabular}
\end{table*}

\subsection{Face reconstruction}
In this part, we demonstrate the performance of the proposed PCA-KMPE algorithm by applying it to the face reconstruction task {\cite{kwak2008principal}}. The Yale face database {\cite{georghiades2001few}} is used, which contains 165 face image. Each image is normalized to 64$\times$64 pixels, and the values of the pixels are set in $[0,255]$. In our experiment, two types of outliers are considered. For the first type, some images are randomly selected, and the selected images are occluded by a rectangular area, where pixels are randomly set at either 0 or 255, and the location of the rectangular area is randomly determined. For the second type, all pixels of the selected images are set at either 0 or 255. Some examples of the first type are illustrated in Fig.2.
The reconstruction performance is measured by the \emph{average reconstruction error}, defined by {\cite{kwak2008principal}}
\begin{equation}
e(m) = \frac{1}{n}\sum\limits_{i = 1}^n {{{\left\| {(\textbf{x}_i^{org} - \boldsymbol{\mu} ) - {\textbf{W}}{{\textbf{W}}^T}({\textbf{x}_i} - \boldsymbol{\mu} )} \right\|}_2}}
\end{equation}
where $\textbf{x}_i^{org}$ and $\textbf{x}_i$ denote, respectively, the original unoccluded image and corresponding training image. For comparison purpose, we also demonstrate the performance of the PCA {\cite{jolliffe2002principal}}, PCA-L1 {\cite{kwak2008principal}}, R1-PCA {\cite{ding2006r}}, PCA-GM R1-PCA {\cite{oh2016generalized}} and HQ-PCA {\cite{he2011robust}}. In the experiment, the kernel widths of the PCA-KMPE and HQ-PCA are selected by (30). The parameter  of HQ-PCA, PCA-GM and PCA-KMPE is set at 10. The average reconstruction errors of the six PCA algorithms versus the number of principal components under two types of outliers are illustrated in Fig.3 and 4. Evidently, the PCA-KMPE algorithm achieves the best performance among all the tested methods.
\begin{figure}[t]
\setlength{\abovecaptionskip}{0pt}
\setlength{\belowcaptionskip}{0pt}
\centering
\includegraphics[width=\linewidth]{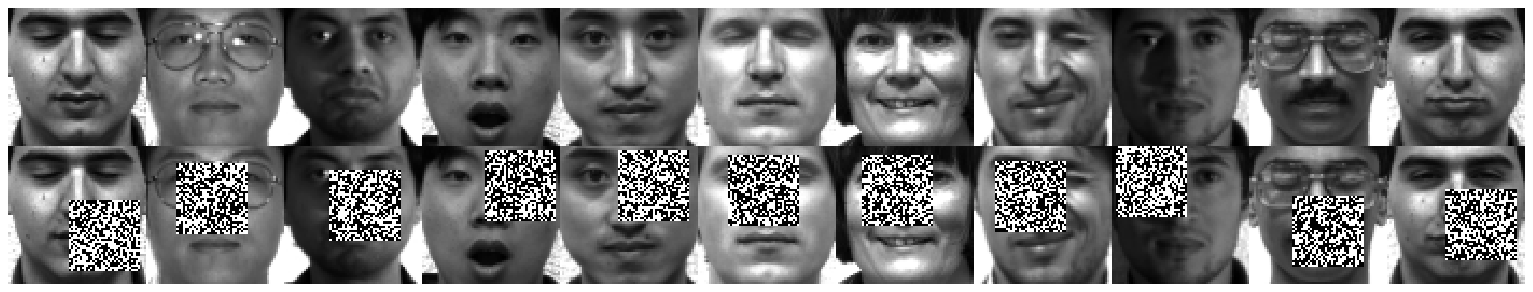}
\caption{Some examples of the original images (upper row) and corresponding contaminated images (low row)}
\label{fig2}
\end{figure}

\begin{figure*}[htbp]
\setlength{\abovecaptionskip}{0pt}
\setlength{\belowcaptionskip}{0pt}
\centering
\subfigure[]{
\includegraphics[width=3.0in,height=2.4in]{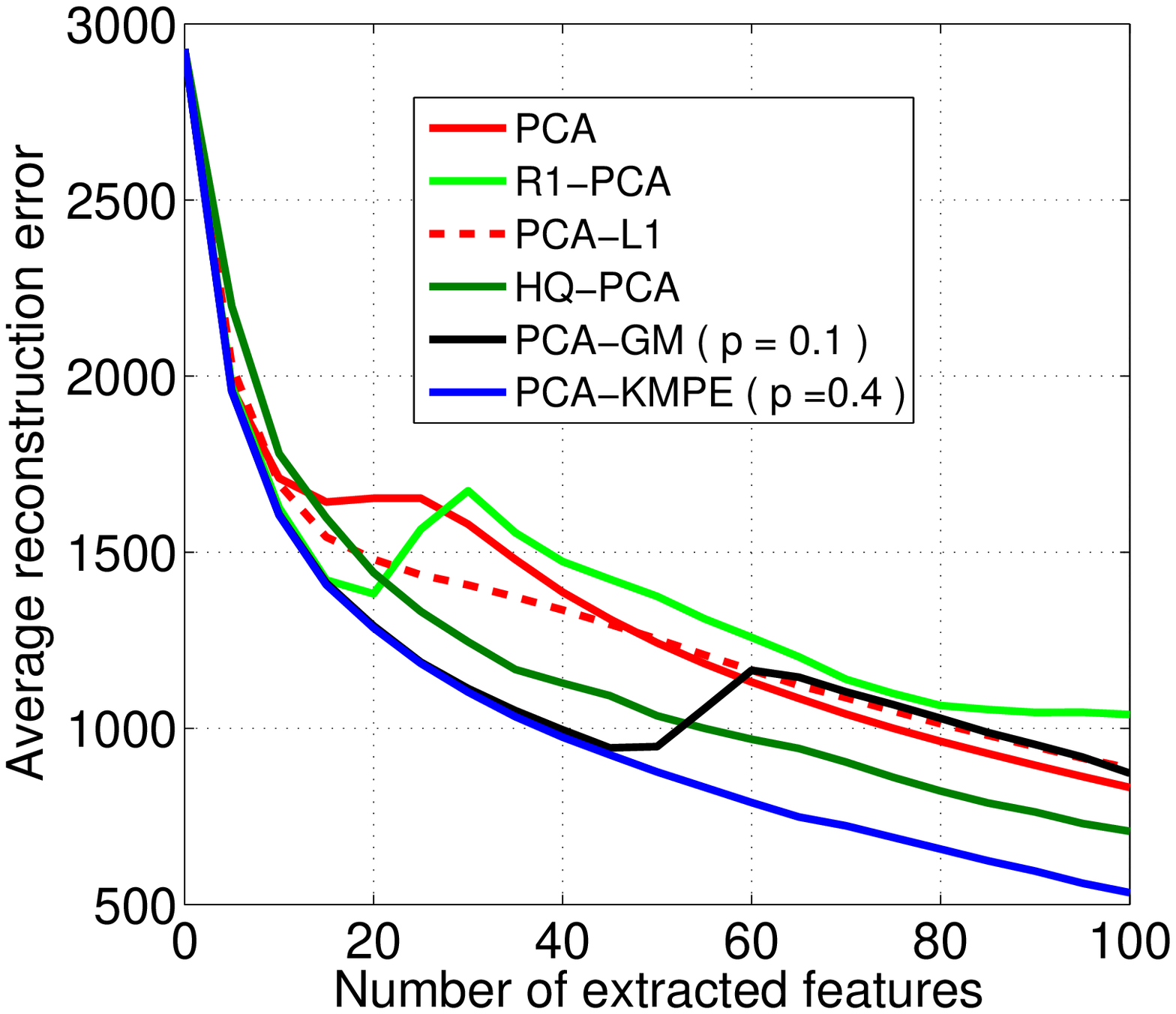}}
\subfigure[]{
\includegraphics[width=3.0in,height=2.4in]{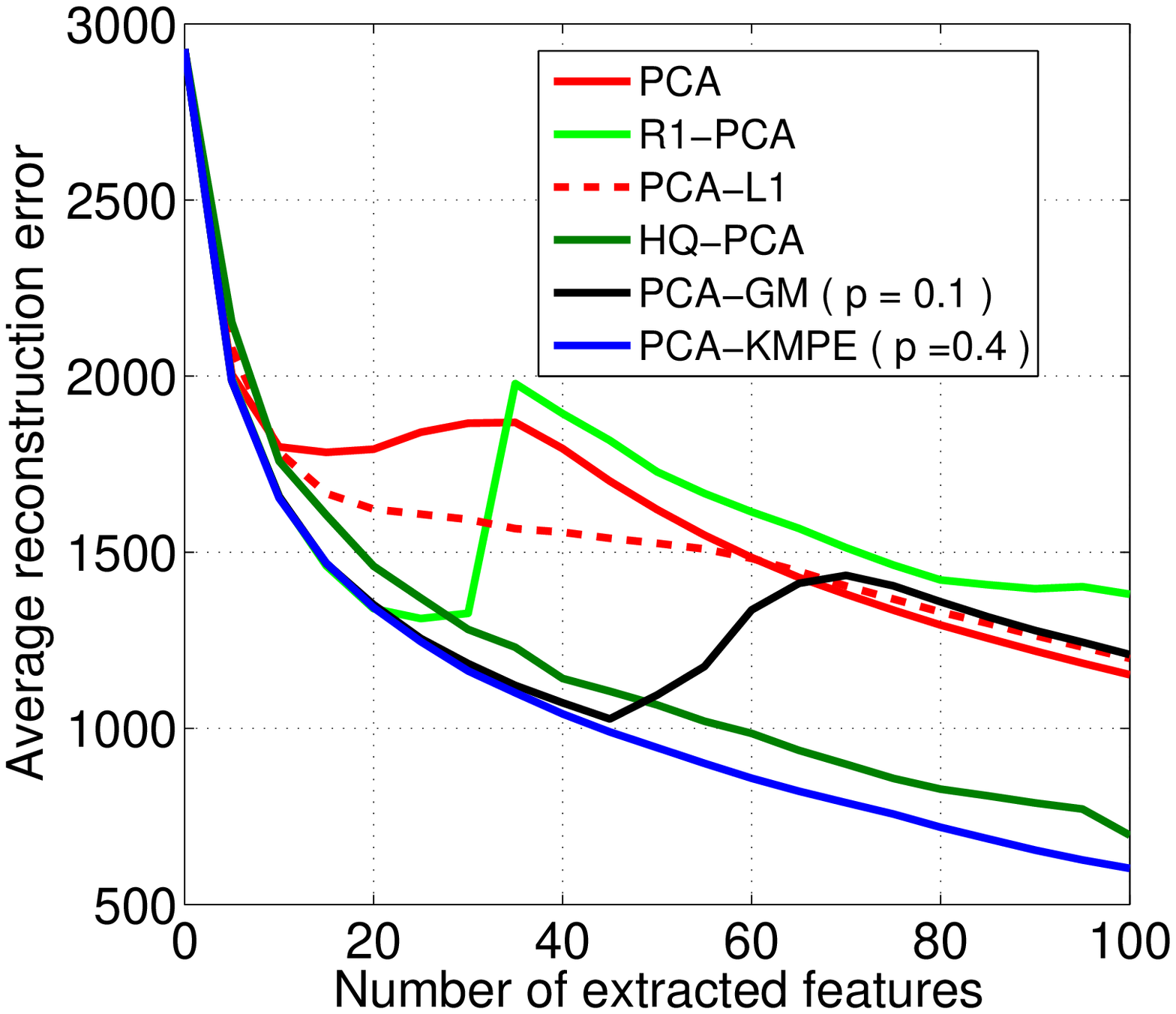}}
\caption{Average reconstruction errors of different PCA algorithms under occlusion images, where the numbers of inliers and outliers are: (a) (150,15); (b) (140,25).}
\label{fig1}
\end{figure*}

\begin{figure*}[htbp]
\setlength{\abovecaptionskip}{0pt}
\setlength{\belowcaptionskip}{0pt}
\centering
\subfigure[]{
\includegraphics[width=3.0in,height=2.4in]{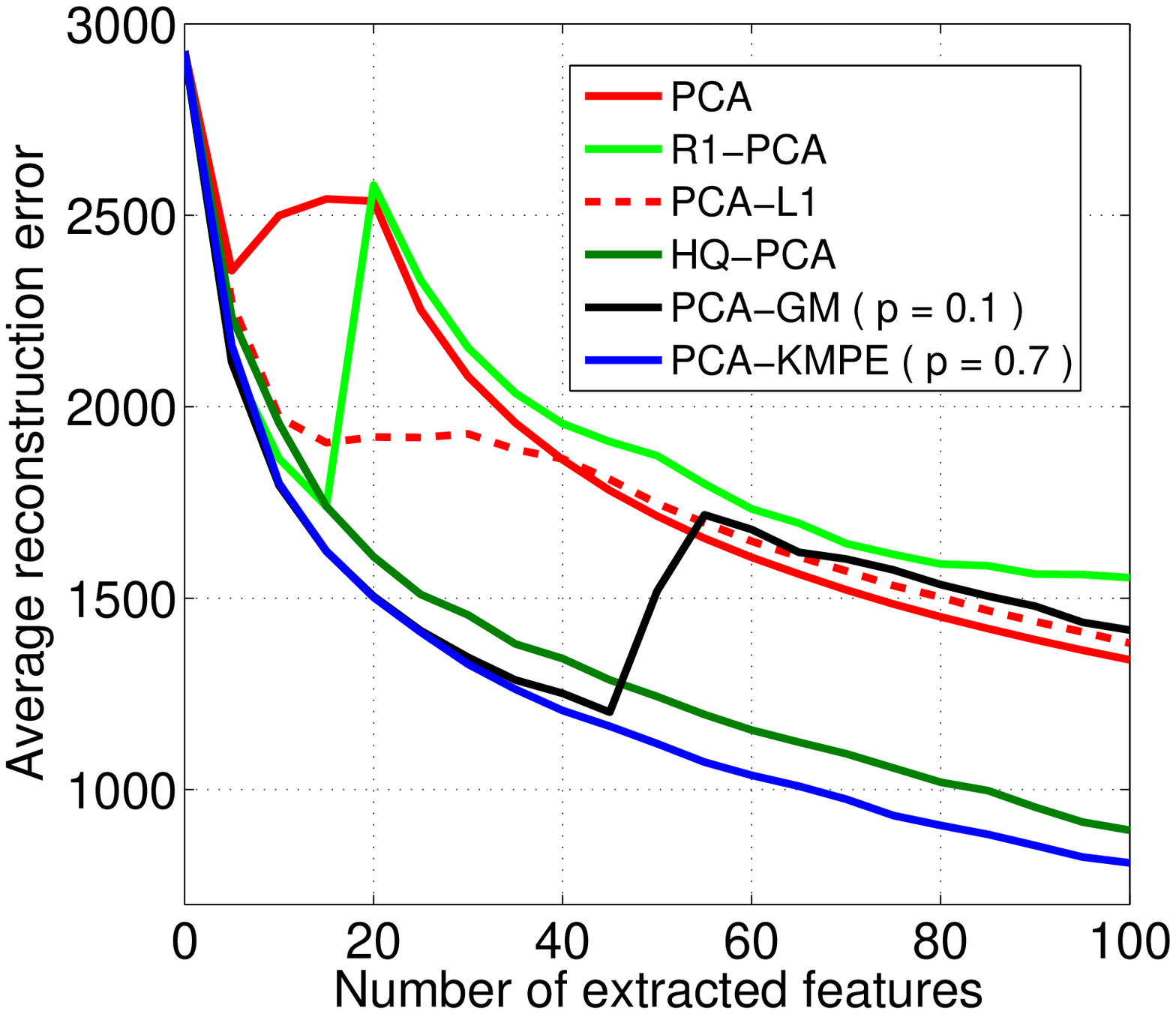}}
\subfigure[]{
\includegraphics[width=3.0in,height=2.4in]{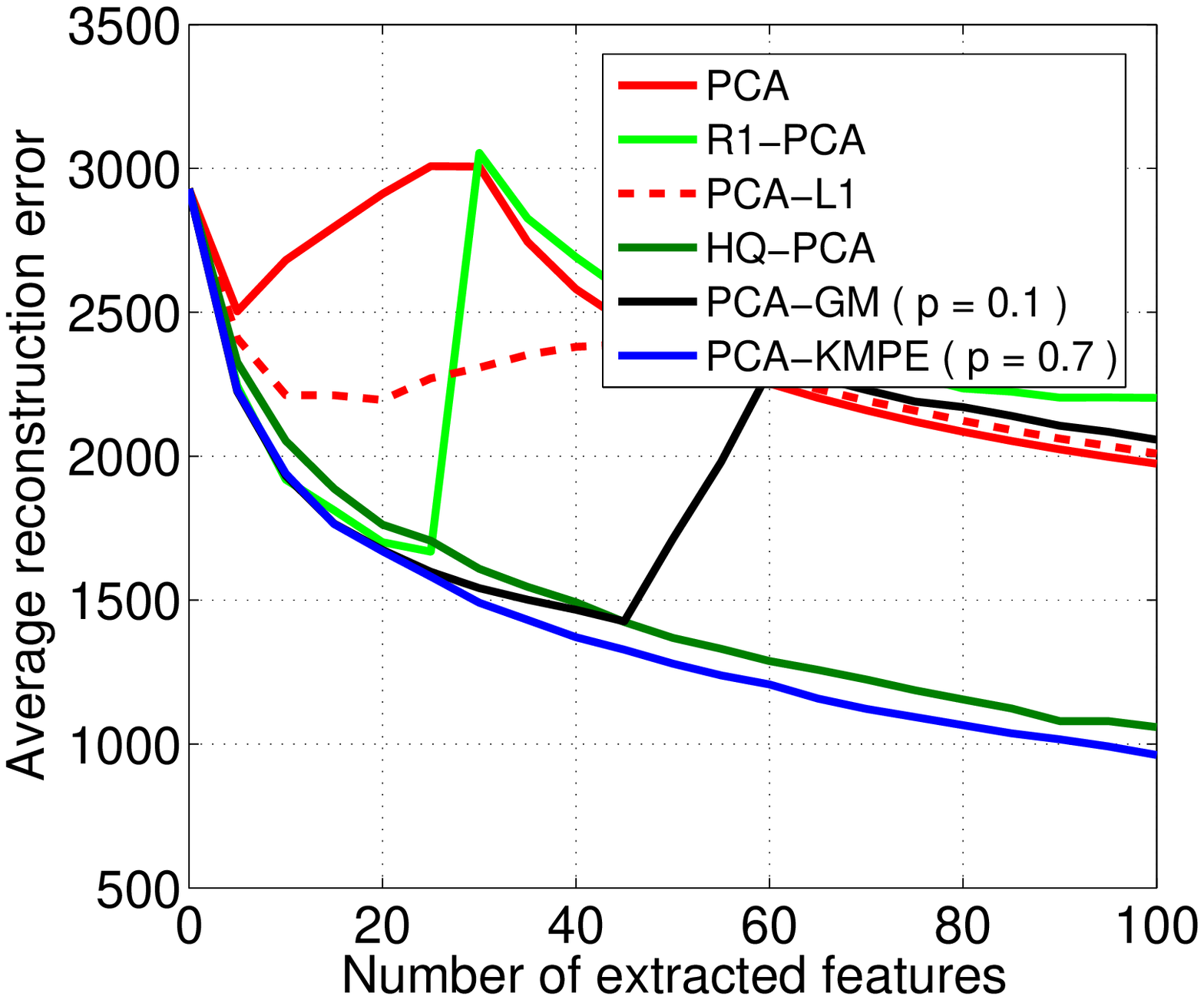}}
\caption{Average reconstruction errors of different PCA algorithms under dummy images, where the numbers of inliers and outliers are: (a) (150, 15); (b) (140, 25).}
\label{fig1}
\end{figure*}

The effectiveness of the proposed PCA-KMPE can also be verified by visualizing the eigenfaces and reconstructed images. The eigenfaces obtained by PCA, R1-PCA, PCA-L1, HQ-PCA, PCA-GM and PCA-KMPE are shown in Fig.5. Due to space limitation, for each method only ten eigenfaces are presented, with $m=10$. In addition, Fig.6 shows the face reconstruction results. These results are achieved under the occlusion and dummy noises (the numbers of the inlier and outlier images are (150, 15)) with the number of the extracted features being 50. Since there are some noisy images (occlusion or dummy) in the training set, most of the eigenfaces (especially those obtained by PCA, R1-PCA and PCA-L1) are contaminated. However, the eigenfaces of PCA-KMPE look very good in visualization. From Fig.6, one can observe that PCA-KMPE can well eliminate the influence by outliers.

\begin{figure*}[htbp]
\setlength{\abovecaptionskip}{0pt}
\setlength{\belowcaptionskip}{0pt}
\centering
\subfigure[]{
\includegraphics[width=3.5in]{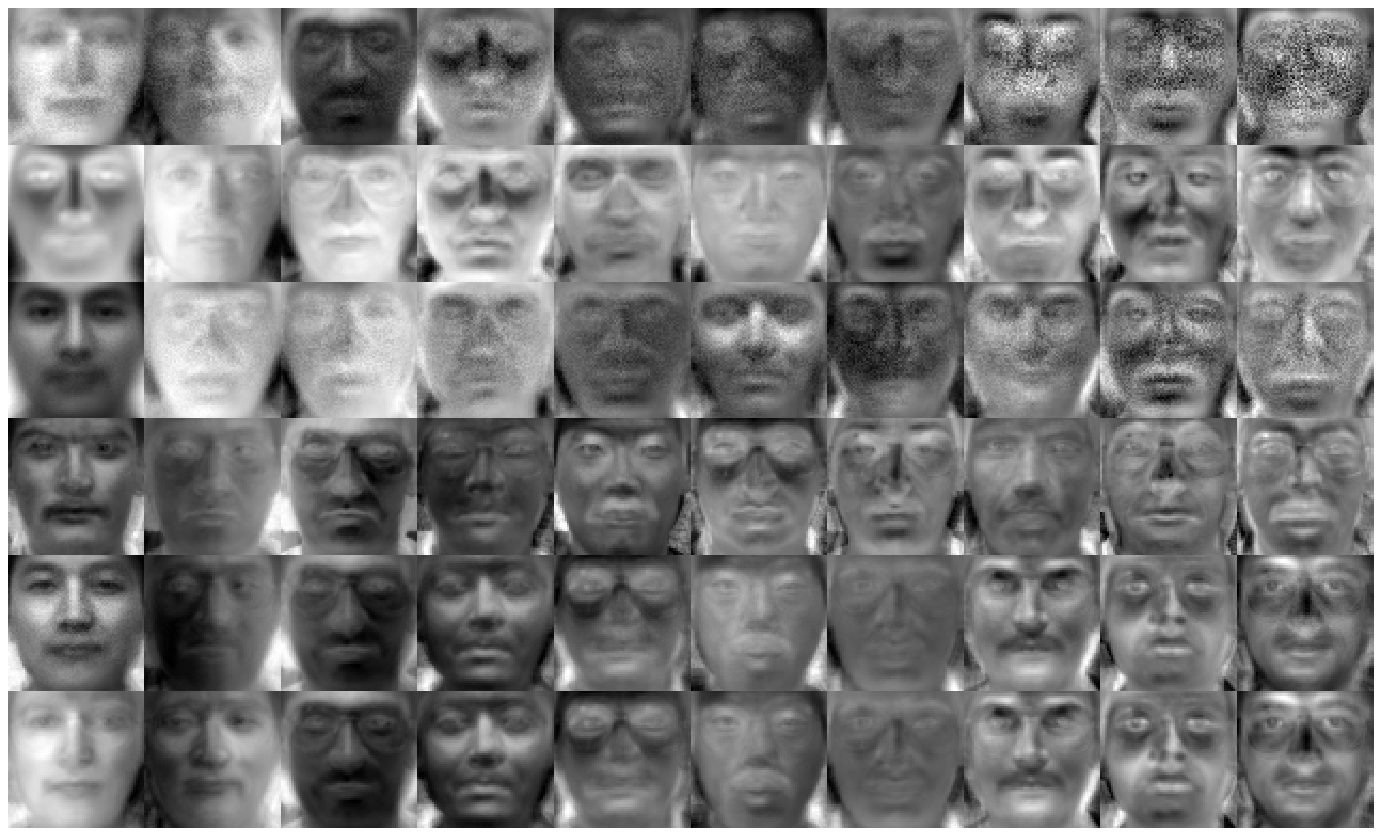}}
\subfigure[]{
\includegraphics[width=3.5in]{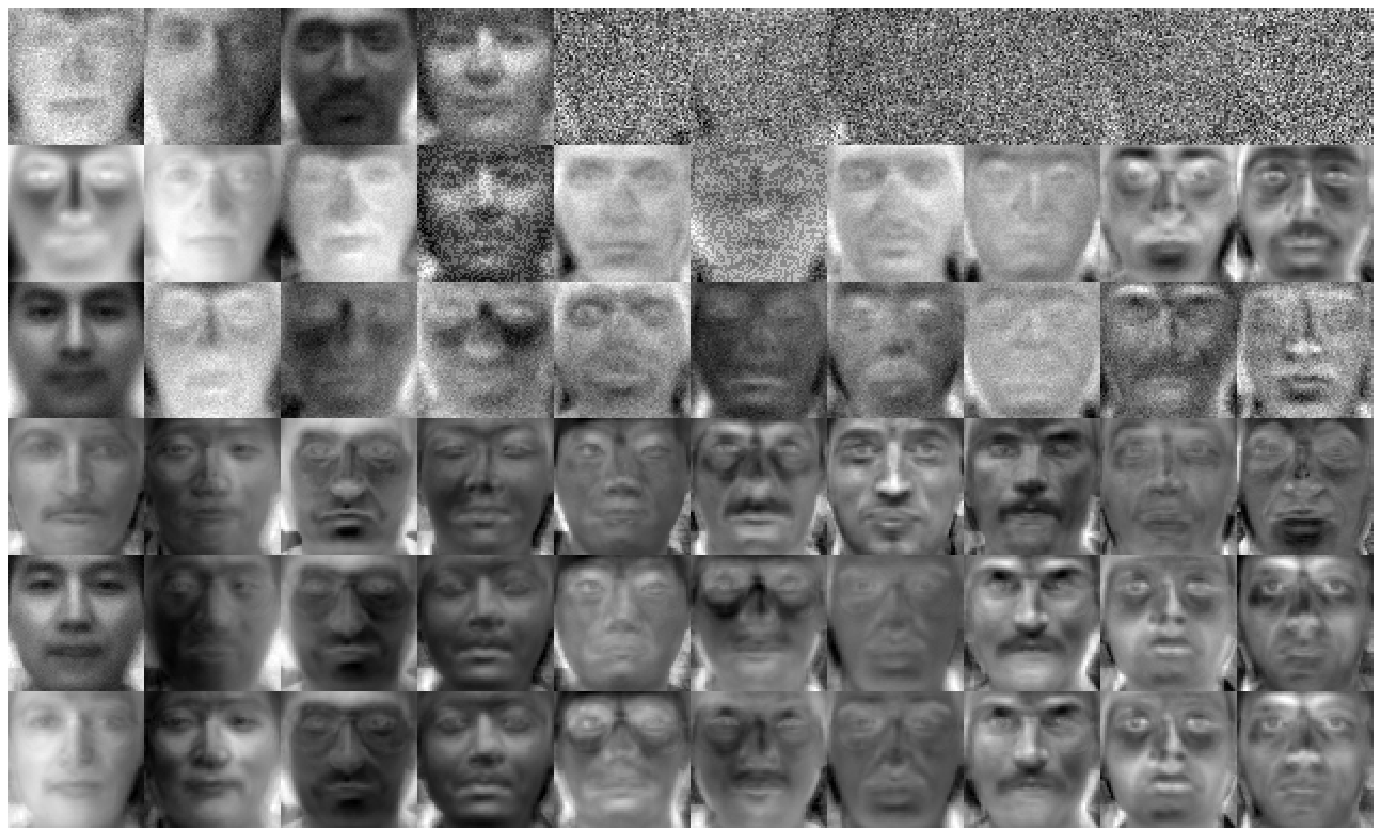}}
\caption{Eigenfaces of six PCA algorithms ($m=10$ ). The images in each row are obtained by PCA, R1-PCA, PCA-L1, HQ-PCA, PCA-GM and PCA-KMPE. (a) occlusion noise; (b) dummy noise}
\label{fig1}
\end{figure*}

\begin{figure*}[htbp]
\setlength{\abovecaptionskip}{0pt}
\setlength{\belowcaptionskip}{0pt}
\centering
\subfigure[]{
\includegraphics[width=3.5in]{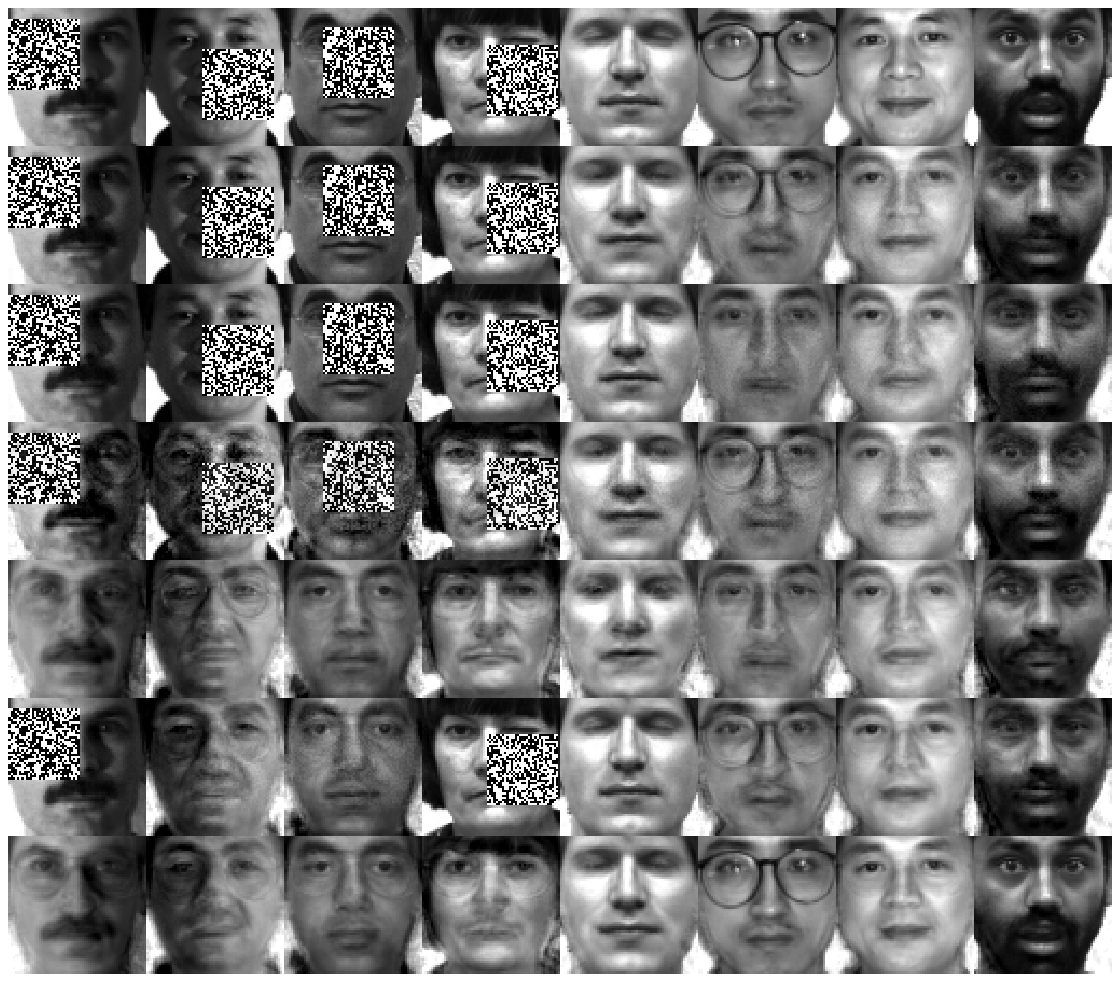}}
\subfigure[]{
\includegraphics[width=3.5in]{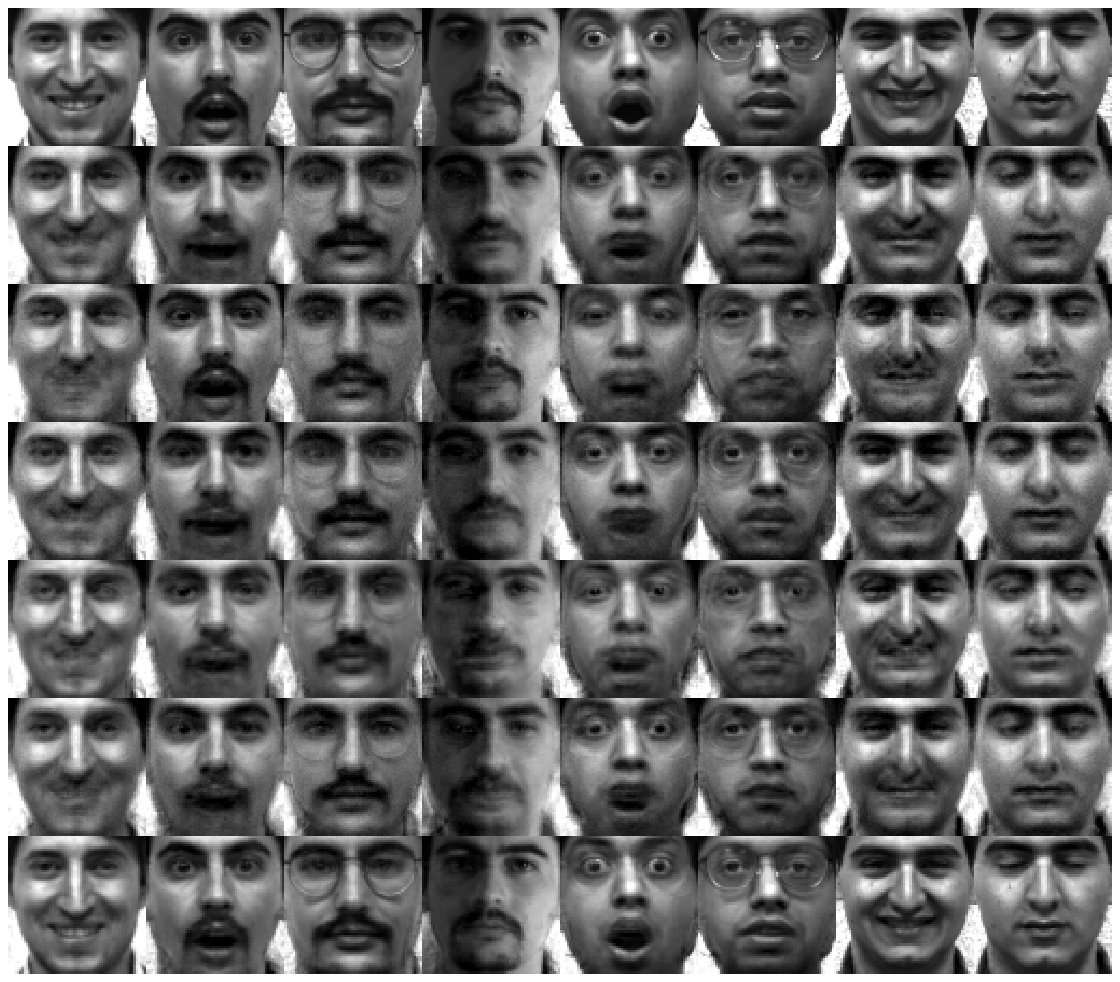}}
\caption{Reconstructed images of six PCA algorithms. The first row shows the training images contaminated with (a) occlusion noise, (b) dummy noise. The rest rows show the images reconstructed by PCA, R1-PCA, PCA-L1, HQ-PCA, PCA-GM and PCA-KMPE.}
\label{fig1}
\end{figure*}

\subsection{Clustering}
Theoretical analysis and experimental results {\cite{he2011robust,ding2006r,oh2016generalized,ding2004k}} in the literature show that PCA methods can be used as a preprocessing step to improve the clustering accuracy of K-means. In the last part, we apply the proposed PCA-KMPE algorithm to a clustering problem with outliers. Two databases, MNIST handwritten digits database and Yale Face database, are chosen in our experiment. The MNIST handwritten digits data contain 60000 samples in training set and 10000 samples in testing set. In the experiment, we randomly select 300 samples of digits $\left\{ {3,{\rm{ }}8,{\rm{ }}9} \right\}$ from the first 10000 samples in training set. Accordingly, 60 samples of other digits as outliers, are selected from the same 10000 samples. Thus the numbers of the outliers and inliers are 60 and 300, respectively. The selected samples are normalized to unit norm before experiment. In Fig.7, the upper and lower rows show the randomly selected inlier and outlier digits from the database. The second database, Yale Face, contains 165 grayscale images of 15 individuals, namely 15 classes. In the experiment, 15 dummy images contaminate the database. The goal is thus to learn a projection matrix from the training data (360 handwritten digital images or 180 faces images) using a PCA method, and obtain the testing results with testing data (noise free) on subspaces. Then we use the K-means algorithm to cluster the PCA results into 3 or 15 classes.

\par We use the clustering accuracy (ACC) and normalized mutual information (NMI) of K-means on subspaces, to quantitatively evaluate the performance of the aforementioned six PCA methods. Let $\emph{\textbf{p}}$ and $\emph{\textbf{t}}$ be the predicted and target label vectors, NMI is defined as
\begin{equation}
NMI = \frac{{I(\emph{\textbf{p}},\emph{\textbf{t}})}}{{\sqrt {H(\emph{\textbf{p}})H(\emph{\textbf{t}})} }}
\end{equation}
where $I(\emph{\textbf{p}},\emph{\textbf{t}})$ is the mutual information between $\emph{\textbf{p}}$ and $\emph{\textbf{t}}$, and $H(\emph{\textbf{p}})$ and $H(\emph{\textbf{t}})$ are the entropies of $\emph{\textbf{p}}$ and $\emph{\textbf{t}}$. Clearly, the higher the values of ACC and NMI, the better the clustering performance. The clustering results on the two databases with different PCA methods are shown in Table 9 and 10, in which the best results under the same dimension number of subspaces are represented in bold. One can see that PCA-KMPE usually achieves the best performance among the six methods.

\begin{figure}[t]
\setlength{\abovecaptionskip}{0pt}
\setlength{\belowcaptionskip}{0pt}
\centering
\includegraphics[width=\linewidth]{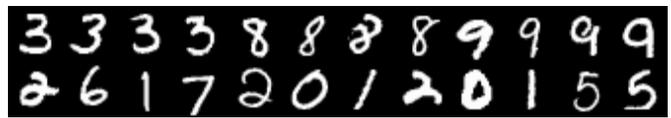}
\caption{Selected digital images from MNIST handwritten database}
\label{fig2}
\end{figure}

\begin{table*}[]\small
\renewcommand\arraystretch{1.5}
\setlength{\abovecaptionskip}{0pt}
\setlength{\belowcaptionskip}{5pt}
\centering
\caption{Clustering accuracy (\%) of the K-means on subspaces of the digital images `3', `8' and `9'}
\begin{tabular}{p{1cm}<{\centering}p{1.7cm}<{\centering}p{1.7cm}<{\centering}p{1.7cm}<{\centering}p{1.7cm}<{\centering}p{2cm}<{\centering}p{2cm}<{\centering}}
\toprule
\multicolumn{1}{c}{m}& \multicolumn{1}{c}{PCA} &\multicolumn{1}{c}{ R1-PCA} &\multicolumn{1}{c}{ L1-PCA }& \multicolumn{1}{c}{HQ-PCA} &\multicolumn{1}{c}{ PCA-GM(p=0.3)} & \multicolumn{1}{c}{PCA-KMPE(p=10)}\\
\hline
 50 &68.22$\pm$6.62	&69.63$\pm$6.32	&63.02$\pm$5.86	&69.92$\pm$6.12	&68.05$\pm$7.23	&\textbf{71.83$\pm$6.06}\\
100	&67.62$\pm$5.87	&67.31$\pm$5.65	&66.31$\pm$5.52	&67.98$\pm$5.77	&67.94$\pm$6.08	&\textbf{69.53$\pm$6.61}\\
150	&68.27$\pm$6.23	&68.56$\pm$5.91	&67.34$\pm$6.44	&69.04$\pm$5.67	&68.38$\pm$5.76	&\textbf{69.55$\pm$5.71}\\
200	&67.87$\pm$6.13	&69.09$\pm$5.72	&68.07$\pm$6.90	&69.35$\pm$6.45	&68.15$\pm$6.24	&\textbf{69.91$\pm$6.85}\\
250	&67.76$\pm$6.45	&67.10$\pm$6.22	&67.34$\pm$5.81	&\textbf{68.62$\pm$6.25}	&67.76$\pm$6.45	&68.57$\pm$6.49\\
300	&67.19$\pm$5.99	&67.15$\pm$6.03	&67.18$\pm$5.99	&67.85$\pm$6.69	&67.18$\pm$5.99	&\textbf{68.03$\pm$6.55}\\
\bottomrule
\end{tabular}
\end{table*}

\begin{table*}[]\small
\renewcommand\arraystretch{1.5}
\setlength{\abovecaptionskip}{0pt}
\setlength{\belowcaptionskip}{5pt}
\centering
\caption{ACC and NMI of the K-means on subspaces of Yale Face database with dummy noise}
\begin{tabular}{p{0.5cm}<{\centering}p{0.5cm}<{\centering}p{2cm}<{\centering}p{2cm}<{\centering}p{2cm}<{\centering}p{2cm}<{\centering}p{2.3cm}<{\centering}p{2.3cm}<{\centering}}
\toprule
\multicolumn{1}{c}{m}&{}& \multicolumn{1}{c}{PCA} &\multicolumn{1}{c}{ R1-PCA} &\multicolumn{1}{c}{ L1-PCA }& \multicolumn{1}{c}{HQ-PCA} &\multicolumn{1}{c}{ PCA-GM(p=0.3)} & \multicolumn{1}{c}{PCA-KMPE(p=10)}\\
\hline
\multirow{2}*{20}&ACC&0.4872$\pm$0.0433&0.4898$\pm$0.0382&0.4886$\pm$0.0380&0.4832$\pm$0.0410&0.4933$\pm$0.0419&\textbf{0.4933$\pm$0.0366}\\
&NMI	&0.5618$\pm$0.0277&	0.5567$\pm$0.0225&	0.5620$\pm$0.0227&	0.5375$\pm$0.0262	&\textbf{0.5637$\pm$0.0256}&	0.5604$\pm$0.0233\\
\multirow{2}*{40}&ACC&0.4809$\pm$0.0462	&0.4848$\pm$0.0365&\textbf{0.4981$\pm$0.0405}&	0.4907$\pm$0.0413&	0.4859$\pm$0.0429	&0.4927$\pm$0.0446\\
&NMI &0.5589$\pm$0.0317	&0.5596$\pm$0.0254	&0.5682$\pm$0.0264	&0.5576$\pm$0.0275	&0.5612$\pm$0.0270	&\textbf{0.5715$\pm$0.0289}\\
\multirow{2}*{60}&ACC&0.4775$\pm$0.0444	&0.4923$\pm$0.0498&	0.4980$\pm$0.0400	&0.4925$\pm$0.0390	&0.4935$\pm$0.0398	&\textbf{0.5133$\pm$0.0401}\\
&NMI	&0.5570$\pm$0.0292	&0.5657$\pm$0.0318&	0.5693$\pm$0.0274&	0.5617$\pm$0.0272&	0.5658$\pm$0.0278&	\textbf{0.5809$\pm$0.0281}\\
\multirow{2}*{80}&ACC&0.4938$\pm$0.0408	&0.4887$\pm$0.0428	&0.4903$\pm$0.0476	&0.4900$\pm$0.0337&	0.4851$\pm$0.0444	&\textbf{0.5020$\pm$0.0432}\\
&NMI&0.5651$\pm$0.0291	&0.5664$\pm$0.0295	&0.5655$\pm$0.0330	&0.5608$\pm$0.0268&	0.5633$\pm$0.0319&	\textbf{0.5741$\pm$0.0284}\\
\multirow{2}*{100}&ACC&0.4856$\pm$0.0430&	0.4801$\pm$0.0422&	0.4891$\pm$0.0489&	0.4892$\pm$0.0423&	0.4812$\pm$0.0497&	\textbf{0.4956$\pm$0.0423}\\
&NMI	&0.5648$\pm$0.0277&	0.5580$\pm$0.0318&	0.5640$\pm$0.0324&	0.5587$\pm$0.0284&	0.5602$\pm$0.0339&	\textbf{0.5694$\pm$0.0292}\\
\bottomrule
\end{tabular}
\end{table*}

\section{CONCLUSION}
A new statistical measure in kernel space is proposed in this work, called the \emph{kernel mean-p power error} (KMPE), which generalizes the \emph{correntropic loss} (C-Loss) to the case of arbitrary power, and some basic properties are presented. In addition, we consider two application examples, \emph{extreme learning machine} (ELM) and \emph{principal component analysis} (PCA), and two robust learning algorithms are developed by using KMPE as loss function, namely ELM-KMPE and PCA-KMPE. Experimental results show that the new algorithms can consistently outperform some existing methods in function estimation, regression, classification, face reconstruction and clustering.

\bibliographystyle{unsrt}
\bibliography{KMPE}
\end{document}